\documentclass{article}
\usepackage{arxiv}

\usepackage{amsmath}
\usepackage[nopatch]{microtype}
\usepackage{booktabs}  
\usepackage{amsthm}
\usepackage{enumerate}%

\usepackage[utf8]{inputenc} 
\usepackage[T1]{fontenc}    
\usepackage{hyperref}       
\usepackage{url}            
\usepackage{booktabs}       
\usepackage{nicefrac}       
\usepackage{microtype}      
\usepackage{graphicx}
\usepackage{natbib}
\usepackage{doi}

\usepackage[thicklines]{cancel}
\theoremstyle{definition}
\newtheorem{definition}{\textit{Assumption}}[section]
\newtheorem*{theorem}{\textit{Theorem}}
\theoremstyle{remark}

\usepackage{multirow}
\usepackage{multicol}

\usepackage{lineno}

\usepackage{tikz}
\newcommand*\circled[1]{\tikz[baseline=(char.base)]{
            \node[shape=circle,draw,inner sep=2pt] (char) {#1};}}

\usepackage{amssymb}
\usepackage{subfigure}

\newtheorem{lemma}{\textit{Lemma}}

\title{Commuting Distance Regularization for Timescale-Dependent Label Inconsistency in EEG Emotion Recognition}

\author{
  Xiaocong Zeng \\
  School of Mathematics (Zhuhai) \\
  Sun Yat-sen University \\
  Guangdong, 519082, CHINA \\
  \texttt{zengxc3@mail2.sysu.edu.cn} \\
    \And
  Craig Michoski \\
  ODEN Institute for Computational Engineering \& Sciences \\
  University of Texas at Austin\\
  Austin, Texas 78712-1229, USA \\
  \texttt{michoski@oden.utexas.edu} \\
  \And
  \href{https://orcid.org/0000-0002-6483-8326}{\includegraphics[scale=0.06]{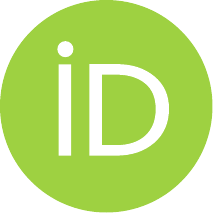}\hspace{1mm}Yan Pang} \\
Shenzhen Institute of Advanced Technology \\
Chinese Academy of Sciences \\
Shenzhen, CHINA\\
\texttt{yanpang@siat.ac.cn} \\
\And
\href{https://orcid.org/0000-0002-4862-7182}{\includegraphics[scale=0.06]{orcid.pdf}\hspace{1mm}Dongyang Kuang*} \\
School of Mathematics (Zhuhai)\\
Sun Yat-sen University\\
Guangdong, 519082, CHINA \\
\texttt{kuangdy@mail.sysu.edu.cn} \\
}

\hypersetup{
pdftitle={Mitigating Timescale-Dependent Label Inconsistency in EEG-based Emotion Recognition via Emotion Transition Graph Regularization},
pdfauthor={Dongyang Kuang, Xiaocong Zeng, Yan Pang and Craig Michoski},
pdfkeywords={Affective Computing, Emotion Recognition, EEG, Label Noise, Variation Regularization, Commuting Distance},
}

\begin{document}
\maketitle

\begin{abstract}
In this work, we address the often-overlooked issue of Timescale Dependent Label Inconsistency (TsDLI) in training neural network models for EEG-based human emotion recognition. To mitigate TsDLI and enhance model generalization and explainability, we propose two novel regularization strategies: Local Variation Loss (LVL) and Local-Global Consistency Loss (LGCL). Both methods incorporate classical mathematical principles—specifically, functions of bounded variation and commute-time distances—within a graph-theoretic framework. Complementing our regularizers, we introduce a suite of new evaluation metrics that better capture the alignment between temporally local predictions and their associated global emotion labels. We validate our approach through comprehensive experiments on two widely used EEG emotion datasets, DREAMER and DEAP, across a range of neural architectures including LSTM and transformer-based models. Performance is assessed using five distinct metrics encompassing both quantitative accuracy and qualitative consistency. Results consistently show that our proposed methods outperform state-of-the-art baselines, delivering superior aggregate performance and offering a principled trade-off between interpretability and predictive power under label inconsistency. Notably, LVL achieves the best aggregate rank across all benchmarked backbones and metrics, while LGCL frequently ranks the second, highlighting the effectiveness of our framework.
\end{abstract}

\noindent 

\section{Introduction}\label{sec:intro}



Electroencephalography (EEG) based emotion recognition is a cutting-edge field that leverages the electrical activity of the brain to identify and classify human emotions. In recent years, the uprising of deep learning methodologies has revolutionized EEG-based affective computing by providing powerful tools to automatically learn and extract complex patterns from either manually extracted features or directly from raw EEG signals \cite{li2022eeg}. Nonetheless, many problems still remain for further and in-depth investigations. This paper will focus on a particular problem that is rarely discussed in the context of EEG based emotion recognition tasks.

We refer to this issue as the \textbf{Timescale-Dependent Label Inconsistency (TsDLI)} problem. TsDLI emerges naturally from the typical experimental protocols employed in EEG-based emotion recognition data collection. Popular open-source datasets such as SEED \cite{duan2013differential,zheng2015investigating}, DEAP \cite{koelstra2011deap}, and DREAMER \cite{katsigiannis2017dreamer} commonly involve presenting multimedia stimuli to subjects for an extended duration, after which participants provide a single global emotion label—either as discrete emotion categories (e.g., happy or sad) or numerical ratings (e.g., valence/arousal scores from 1 to 5)—that summarizes their overall experience. 

\begin{figure}
    \centering
    \includegraphics[width=\textwidth]{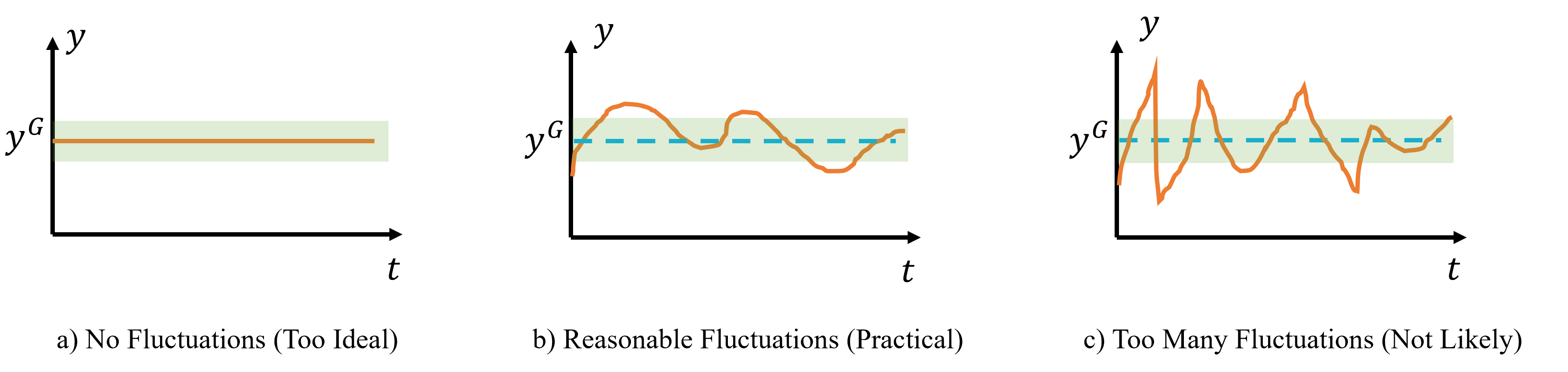}
    \caption{A visual representation of the \textit{Timescale Dependent Label Inconsistency (TsDLI)} problem. a) The assumption that actual emotion track $y(t)$ agrees with the self-reported global emotion label $y^G$ is too ideal to be true. Case b) where $y(t)$ during experiment fluctuates reasonably around $y^G$ is more practical and logical. Too much fluctuations such as case c) is not likely to happen either, especially when data are collected in Lab environments. }
    \label{fig:TsDLI}
\end{figure}

In contrast, many practical applications, particularly emotion recognition modules integrated into consumer-grade devices such as earbuds or VR/AR headsets, require real-time or near-real-time emotion predictions based on short-duration EEG segments. To accommodate such constraints, previous studies \cite{li2022eeg,cui2020eeg,liu2024vbh} often partition continuous EEG recordings into shorter segments and directly inherit the global label reported by the subject as the local label for each segment. However, this simplistic label assignment strategy neglects the inherently dynamic and temporally fluctuating nature of human emotional states \cite{waugh2015temporal,puccetti2022temporal,gutentag2022incremental}. As illustrated in Figure~\ref{fig:TsDLI}, the emotion labels that reflect short-term, moment-to-moment variations can substantially differ from the self-reported global labels because they are conditioned on distinct timescales. Such timescale-dependent inconsistencies pose significant challenges to accurate pattern learning at short timescales. Moreover, explicitly annotating short segments is highly challenging and often infeasible, especially since most existing datasets lack such fine-grained temporal annotation.

To effectively address the timescale-dependent label inconsistency (TsDLI) problem and thereby develop more reliable and robust EEG-based emotion recognition models, at least three key components must be carefully established. Specifically, these critical components include:

\begin{itemize}
    \item[\textbf{C.1}] \textbf{Emotion Transition Modeling:}  
    A well-defined model for accurately capturing and describing local emotion transitions occurring between successive EEG segments, reflecting realistic emotional dynamics.
    
    \item[\textbf{C.2}] \textbf{Regularization of Label Fluctuations:}  
    A principled strategy to control and regularize the modeled fluctuations, ensuring coherence both between adjacent local predictions and between local predictions and the global emotion label associated with the entire trial.
    
    \item[\textbf{C.3}] \textbf{Comprehensive Performance Evaluation Metrics:}  
    A set of metrics enabling a thorough quantitative and qualitative evaluation of the model's performance, specifically designed to capture the model’s effectiveness in managing TsDLI-related issues.
\end{itemize}


To address the TsDLI problem, we propose two novel loss functions: the \textit{Local Variation Loss (LVL)} and the \textit{Local-Global Consistency Loss (LGCL)}. These losses provide explicit regularization during model training, effectively guiding the learning process to handle potential inconsistencies arising from discrepancies between local emotion predictions and global emotion labels.

Our contributions addressing this issue are summarized as follows:
\begin{itemize}
    \item We formally identify and analyze the Timescale-Dependent Label Inconsistency (TsDLI) problem, an important but understudied issue in EEG-based emotion recognition tasks.
    \item We propose two novel loss functions (Local Variation Loss and Local-Global Consistency Loss) that serve as additional regularization terms, enabling more robust and consistent model training despite the presence of potentially inconsistent labels across short EEG segments.
    \item We introduce several new metrics designed to qualitatively assess prediction consistency, particularly useful in scenarios where explicit local labels for short EEG segments are unavailable.
\end{itemize}



\section{Related Work}\label{sec:related}

TsDLI has received comparatively little attention in the context of EEG-based emotion classification. Nevertheless, it can be conceptualized within the broader framework of learning with noisy labels (LNL) \citep{song2022learning}, wherein discrepancies between local labels and a global label $y^G$
resemble label noise. Even so, TsDLI presents additional complexities that distinguish it from standard LNL settings. In particular, TsDLI employs both local and global labeling scales, while many LNL methods assume all labels reside on a single scale. Furthermore, local labels in TsDLI are often impossible to verify through additional human annotation, unlike some LNL techniques that rely on a small or partially pristine labeled set for validation. Finally, because TsDLI lacks a fully clean test set, it necessitates indirect evaluations rather than conventional accuracy-based metrics. Although certain LNL strategies do not strictly require verified ground truth, TsDLI's reliance on multi-scale labels and unverified local annotations nevertheless poses unique challenges within the noisy-label paradigm


Given these considerations, existing methods in LNL may be adapted to address TsDLI in EEG-based settings. Some representative state-of-the-art LNL methods\footnote{Recently, some studies have started to explore robust methods for other data types (such as time series and graph data) \cite{ma2023ctw,liu2023scale,wang2025noisygl}, but there is still limited research on EEG-based emotion recognition.} are outlined below:

\begin{itemize}
    \item \textbf{Robust Loss Function Approaches:} \cite{ghosh2017robust} established a sufficient condition for robust loss functions and demonstrated the resilience of the Mean Absolute Error (MAE) loss. However, the MAE loss tends to underperform when dealing with complex data. Hence, the Generalized Cross Entropy (GCE) loss of \cite{zhang2018generalized} was introduced to integrate the strengths of MAE and CCE, offering a more flexible noise-robust loss that encompasses both. Drawing on the symmetricity of the Kullback-Leibler divergence, \cite{wang2019symmetric} proposed the Symmetric Cross Entropy (SCE), which combines a noise-tolerant reverse cross-entropy term with the standard cross-entropy loss. Recently, StudentLoss~\cite{zhang2024student} introduces a robust loss function by assuming that samples sharing the same label follow a common student distribution. It is naturally data-selective and offers additional robustness against mislabeled samples.
    \item \textbf{Sample Selection Approaches:} These approaches reduce the influence of noisy labels by identifying clean samples. Exploiting the memory effect of DNNs, they consider examples with small loss values as reliable during training. \cite{han2018co} introduced Co-teaching, which employs two networks that exchange selected samples for parameter updates. Since noisy sample predictions can fluctuate during early training, the SELF found in \cite{nguyen2019self} uses an exponential moving average to stabilize both the network and its predictions for sample selection. The DivideMix approach presented in \cite{li2020dividemix} builds on Co-teaching and semi-supervised learning by integrating a semi-supervised framework called MixMatch, discussed in  \cite{berthelot2019mixmatch}, to treat all noisy samples as unlabeled data and exploit them more effectively.
    \item \textbf{Label Correction (Refinement) Methods:} 
    Instead of discarding samples (as seen in sample selection approaches) presumed to be noisy, label correction approaches attempt to refine or correct the labels themselves during training. \citet{tanaka2018joint} proposed a joint optimization framework, where both network parameters and label estimates are updated iteratively, allowing the model to self-correct noisy labels. Such methods can be combined with consistency regularization \citep{tarvainen2017mean} or data augmentation \citep{cubuk2019autoaugment}, further stabilizing label refinement throughout the training process.
    \item \textbf{Meta-Learning-Based Approaches:}
    An emerging line of work employs meta-learning to discover how to downweight, or correct, noisy samples. \cite{ren2018learning} proposed a method that uses a small set of trusted data to guide the reweighting of training examples in the presence of noise. By treating the weight assignment as a meta-step, the approach dynamically adapts during training and alleviates the adverse effects of label noise. Similar meta-learning strategies have been extended to other problem settings, highlighting their potential adaptability to TsDLI challenges.
    \item \textbf{Transition Matrix Estimation Approaches.}
    Instead of discarding or correcting noisy labels directly, these methods attempt to model how labels are corrupted, often via a transition matrix. Patrini et al.~\citep{patrini2017making} proposed estimating a noise transition matrix, then applying a loss correction to compensate for label corruption. Similar ideas have been extended to multi-class settings~\citep{song2021multiclass} and scenarios where data are highly imbalanced~\citep{buda2018systematic}, underscoring the importance of accurately modeling the noise process for robust learning in more complex label configurations.

    \item \textbf{Distillation/Teacher-Student Approaches.}
    Knowledge distillation methods offer a teacher-student paradigm for training models in the presence of noisy data. A teacher model, often trained on comparatively cleaner or augmented data, transfers its knowledge to a student model, which learns refined representations and predictions. \citet{li2021learning} introduce a strategy wherein a high-confidence teacher filters out, or re-labels, noisy samples before guiding the student, thereby reducing the student model's exposure to label errors. Distillation-based techniques have also proven effective in semi-supervised pipelines, where unlabeled data are leveraged alongside teacher predictions \citep{xie2020self,furlanello2018born}. By integrating knowledge from partially clean or augmented sets, these teacher-student frameworks can help mitigate overfitting to noisy labels while bolstering the overall robustness and generalization of the final trained model.

\end{itemize}

When these LNL techniques are benchmarked for the proposed TsDLI problem, the directly inherited global label $y^G$ is utilized as the local (and potentially noisy) label for each short time signal segment.

\section{Proposed Methods}\label{sec:method}



Although the concept of functions of bounded variation from mathematics offers a straightforward framework for modeling fluctuations in emotion labels, it is not directly applicable to classification tasks that utilize one-hot encoded labels. On one hand, traditional measures of function variation---such as the Mean Absolute Error (MAE)---are computed using model outputs as real numbers, yet numerous studies have demonstrated that this approach is less effective in classification tasks where one-hot encoding is paired with a softmax loss \cite{goodfellow2016deep,janocha2017loss,wang2020comprehensive}. On the other hand, while one-hot encoding is highly effective for standard classification tasks, the Euclidean difference between two one-hot vectors does not adequately capture the true magnitude of fluctuations between their associated emotional states due to its isotropic nature. For example, the fluctuation is identical when the emotion score rises from 1 to 2 or from 1 to 3:
\[ ||(1,0,0) - (0,1,0)||_{l2} = ||(1,0,0) - (0,0,1)||_{l2} = \sqrt{2}, \]
even though the latter change should intuitively represent a larger fluctuation. To reconcile these conflicting aspects and to further incorporate potential non-homogeneous emotion transition priors, we propose two novel loss functions that leverage insights from random walks on a graph.


\subsection{Random Walk on Emotion Transition Graph}

\begin{figure}
    \centering
    \includegraphics[width=0.75\textwidth]{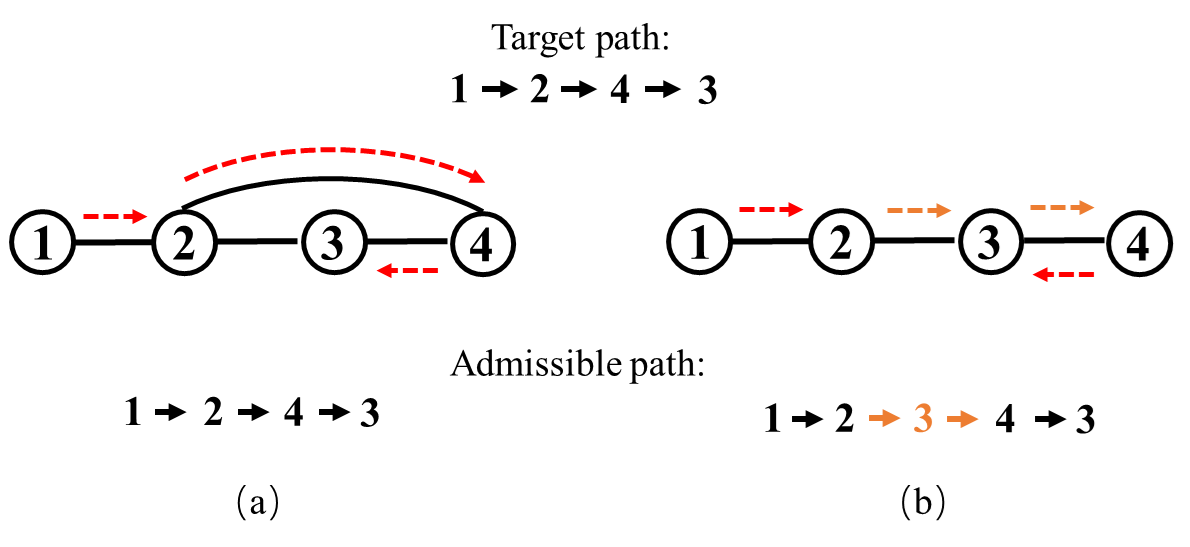}
    \caption{Depending on the graph $G(V,E)$, the target path may be a subset of actual admissible path. Case(a) shows a graph where the target path can be achieved with the same length. Case (b) is an example of the shortest admissible path where the sub-path $2\rightarrow4$ has to be completed via $3$ since there is no direct edge between them.}
    \label{fig:path}
\end{figure}

We propose modeling the aforementioned fluctuation along the “track of emotion” as a random walk on an undirected, emotion-level graph $G(V,E)$. The vertices $V$ correspond to distinct valence (arousal) levels, and the edges $E$ describe possible transitions between these levels. For example, if the emotion track over time is $1 \rightarrow 2 \rightarrow 4 \rightarrow 3$, then the corresponding random walk on the graph is $V_1 \rightarrow V_2 \rightarrow V_4 \rightarrow V_3$, assuming $G(V,E)$ is connected. From a modeling perspective, the structure of $G(V,E)$ not only governs admissible local emotion transitions but also allows for the incorporation of logical priors or assumptions (see Figure~\ref{fig:path} for a simple example).  This random walk framework addresses \textbf{C.1} by offering a suitable model for local emotion transitions. Furthermore, it also enables the use of ``commute distance'' (or `resistance distance') as a measure of the extent of probable emotion fluctuation between levels $i$ and $j$, thereby helping address the other key component \textbf{C.2}.


By definition, the commute distance (or resistance distance), $c_{ij}$, between two vertices $v_i$ and $v_j$ is the expected time for a random walk to travel from $v_i$ to $v_j$ and back \cite{lovasz1993random,von2007tutorial}. In contrast to methods that focus on a single shortest path, commute distance considers an ensemble of short paths between collections of vertices. This concept has many appealing properties for studying random walks on graphs and related machine learning methods. One of the earliest classical results relating commute distance and the graph Laplacian appeared in \cite{Klein1993} and is stated as the following theorem.
  \begin{theorem}[Commute Time Distance]\label{Thm:CD}
    Let $G = (V, E)$ a connected, undirected graph. Denote by $c_{ij}$ the commute distance between vertex $v_i$ and vertex $v_j$, by $e_i$ the one-hot vector whose $i$th entry is 1 and by $L^{\dagger}$ the generalized inverse (or pseudoinverse) of $L$. Then we have $c_{ij} = vol(V )(e_i-e_j)^T L^{{\dagger}}(e_i-e_j)$. 
\end{theorem}

This theorem offers a direct way to compute the commute distance between any two vertices on a graph via the graph Laplacian $L$. By definition, for an undirected graph $G = (V, E)$ with adjacency matrix $A$ (indicating edge connections between vertices) and diagonal degree matrix $D$ (storing each vertex’s degree on the diagonal), the graph Laplacian is given by $L = D - A$. A larger commute distance indicates a greater challenge in transitioning between the two vertices. Given a predefined graph $G(V,E)$ derived from clinical priors for modeling possible local emotion state transitions, we can compute these commute distances to quantify emotion fluctuations across the entire trial.


\subsection{Prior Assumptions on Emotion Transition}\label{sec:PAET}
In developing our approach, we incorporate two fundamental assumptions informed by clinical insights. These assumptions, which we term the \emph{Transition Difficulty Postulate (TDP)} and the \emph{Intermediate Value Assumption (IVA)}, guide the dimensional-type (or continuous-type) emotional dynamics along an axis, and are defined formally below.

\begin{definition}[\emph{Transition Difficulty Postulate (TDP)}]\label{asp:TD}
We posit that transitions between emotional levels become less likely as their numerical distance increases. Formally, for any two emotional states $i$ and $j$, the probability (or ease) of transitioning from $i$ to $j$ decreases as the distance $\lvert i - j\rvert$ grows.
\end{definition}

\begin{definition}[\emph{Intermediate Value Assumption (IVA)}]\label{asp:IVA}
Let $t_a < t_b$ be two time points such that $y(t_a) = i$ and $y(t_b) = j$. Suppose $k$ is an emotional level satisfying $k \in [i, j]$ or $k \in [j, i]$. Then there must exist a time $t_c \in [t_a, t_b]$ for which $y(t_c) = k$. In other words, when an emotion transitions from one level to another, all intermediate levels will appear at least once during the transition interval.
\end{definition}

Roughly speaking, \emph{Assumption~\ref{asp:TD}} is a way to quantitatively modeling the established notion of \emph{emotional inertia} \cite{kuppens2017emotion}. Accompanying this inertial assumption, \emph{Assumption~\ref{asp:IVA}} is a natural extension of the continuity assumption in mathematics, this assumption is also supported by clinical observations that emotional transitions are not abrupt but rather gradual, with intermediate levels being traversed during the transition process \cite{davidson1998affective,scherer2009dynamic}. In dimensional-type emotion models, these assumptions align with clinically observed phenomena by restricting large, abrupt transitions across the emotional continuum and ensuring that intermediate levels are traversed with greater likelihood.




As an illustration of \emph{Assumption~\ref{asp:IVA}}, consider a transition of the valence level from 1 to 3. In this case, the process must pass through the intermediate level 2, even if only briefly.\footnote{This requirement can be viewed as a direct corollary of continuity in mathematics; however, we adopt it here in lieu of imposing a stronger continuity condition.} Based on this assumption, we represent the valence levels as an undirected line graph:
\[
\circled{1} - \circled{2} - \circled{3} - \circled{4} - \circled{5},
\]
thus embedding these assumptions into our random-walk framework. This choice also aligns with the observation that each rating level may not carry the same conceptual weight in self-assessment. For instance, if a participant provides a rating of 3 at a given time, it is typically more likely that they would adjust their rating to 2 or 4, rather than jump directly to 1 or 5, upon (potentially semi-continuous) reassessment. Moreover, ratings 1 and 5 represent more ``extreme'' values (i.e., the endpoints in the line graph), and people often avoid such extremes when intermediate, or ``milder'' levels (2, 3, or 4) are available. Such behaviors also coincide with the classically well-established observation of ``\textit{central tendency bias}" \cite{tourangeau2000psychology,paulhus2007self,saris2014design,douven2018bayesian} in survey and rating scale responses.


\subsection{Local Variation Loss (LVL)}\label{ss:LVL}
 


Building on the above considerations, we introduce a local variational loss defined as:
\begin{equation}\label{eqn:variation}
\text{\emph{Local Variation Loss (LVL):} } L 
= \frac{1}{N} \sum\limits_{i=1}^N 
\bigl(\hat{Y}(t_i) - \hat{Y}(t_{i-1})\bigr)^T 
\, L^\dagger \,
\bigl(\hat{Y}(t_i) - \hat{Y}(t_{i-1})\bigr),
\end{equation}
which serves to regulate the total commute distance during emotional fluctuations. In the above formulation, $\hat{Y}(t_i)$ is the one-hot vector, with its $i$-th entry indicating the model’s confidence in the $i$-th level.


\subsection{Local-Global Consistency Loss (LGCL)}\label{ss:LGCL}
To ensure that globally self-assessed emotion labels represent a coherent summary of locally predicted labels from short segments (with minimal uncertainty), we adopt the following assumption.

\begin{definition}[\emph{Full Expectation Assumption (FEA)}]\label{asp:FEA}
The label $Y_G$ for an entire trial is an overall assessment expressed via an expectation: 
\[
    Y_G = \int_t p(t) Y(t) dt = E_t [Y(t)],
\]
where $p(t)$ is a density function describing how local emotions contribute to the overall self-assessment.
\end{definition}

In practice, if $Y(t)$ is estimated by $E[\hat{Y}(t)]$ and $Y_G$ is estimated by $E[\hat{Y}_G]$, the assumption follows directly from the \emph{Law of Total Expectation}: 
\[
    E[\hat{Y}_G] = E_t\bigl[E(\hat{Y}\mid t=t_i)\bigr].
\]
Given each local prediction $\hat{Y}(t)$, the uncertainty in estimating $Y_G$ as $E_t[Y(t)]$ can be expressed under \textbf{\textit{Assumption \ref{asp:FEA}}} as
\begin{equation}\label{eqn:var}
    V_{\hat{Y}} := \int_t p(t)\Bigl(\hat{Y}(t)-E[Y(t)]\Bigr)^2 dt.
\end{equation}

Combining this assumption with the concept of a graph Laplacian, we propose the following variance loss for regularization during training:
\begin{equation}\label{eqn:variance}
    \text{\emph{Local-Global Consistency Loss (LGCL)}: } \quad
    L = \frac{1}{N}\sum_{i=1}^N \Bigl( \hat{Y}(t_i) - E[\hat{Y}(t)]\Bigr)^T 
    L^\dagger \Bigl(\hat{Y}(t_i) - E[\hat{Y}(t)]\Bigr).
\end{equation}

This variance loss is designed to control the uncertainty such that the global rating, summarized from all segments along a given emotion trajectory, remains consistent with the overall self-assessment. It is derived from Equation~\ref{eqn:var} by assuming a uniform density function $p(t)$ over time and introducing $L^\dagger$ as a kernel in place of the standard identity matrix, thus accounting for label inhomogeneity.

Although motivated differently and intended for distinct regularization purposes, we can demonstrate the equivalence of these two losses (viz. Section \ref{ss:LGCL} and Section \ref{ss:LVL}) under certain conditions (i.e., see the Appendix for details). Our experiments show that these proposed losses both effectively regularize the training process and improve model performance in the TsDLI setting.

\subsection{Evaluation Metrics}\label{sec:eval}

In the presence of TsDLI and scenarios where temporally localized emotion labels are not directly available for evaluation, additional metrics are required to achieve a more comprehensive assessment. We introduce the following novel metrics (\textbf{C.3.}) to address this need.

\begin{figure}
    \centering
    \includegraphics[width=\textwidth]{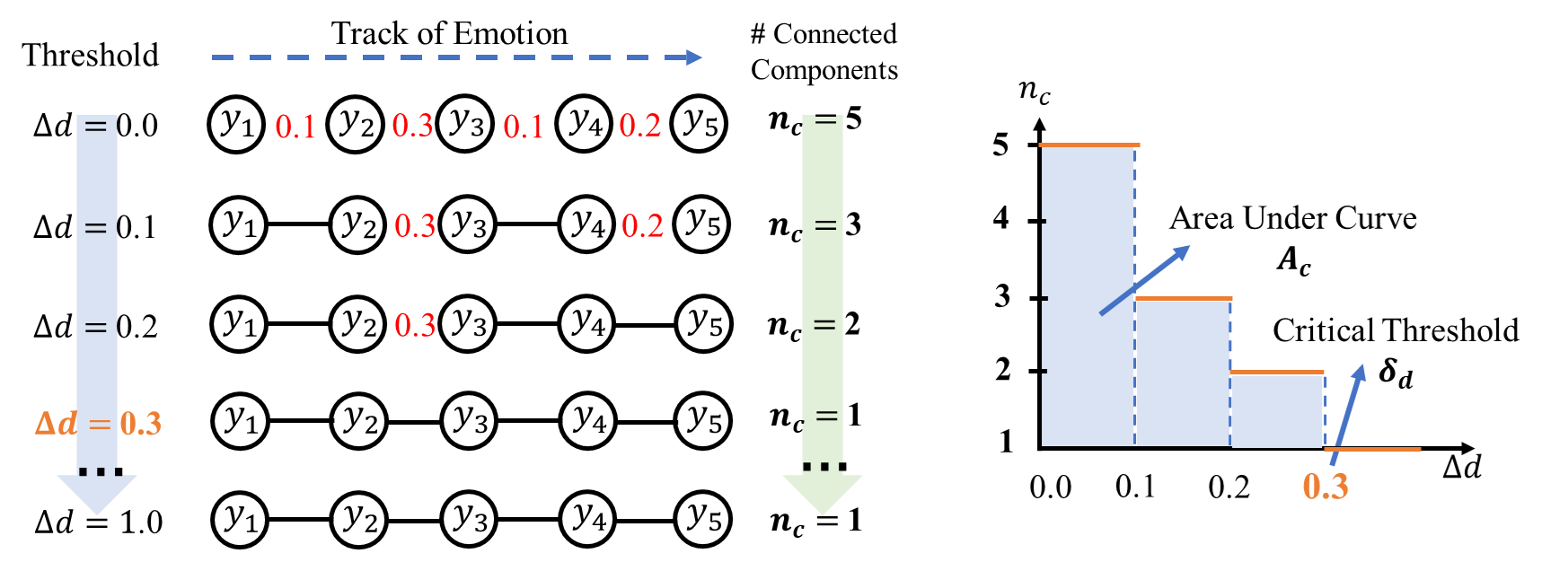}
    \caption{An illustration of the proposed metric is presented. On the left, two consecutive predictions are merged if their distance (shown in red) does not exceed the threshold $\Delta d$, which reduces the number of connected components $n_c$. On the right, $n_c$ is depicted as a non-increasing step function of $\Delta d$, with the limits $\lim_{\Delta d \to 0^+} n_c = n$ and $\lim_{\Delta d \to +\infty} n_c = 1$, where $n$ denotes the total number of signal segments along an emotion track. Notably, a smaller area under the curve---or equivalently, a lower critical threshold at which $n_c$ first becomes 1---indicates that the local segments rapidly converge and align with the global true label.}
    \label{fig:clustermetric}
\end{figure}

Consider predictions on consecutive local segments over time, denoted by $\tau = \{y_1, y_2, \dots, y_n\}$. We merge two consecutive predicted labels, $y_i$ and $y_{i+1}$, if their difference (or "distance") is at most a specified threshold $\Delta d$. Intuitively, smaller differences imply that these segments represent a stable emotional state, whereas larger differences suggest rapid and possibly unrealistic emotional fluctuations.

As $\Delta d$ increases, we become more tolerant of differences between consecutive predictions, leading to fewer separate emotional segments. Hence, the number of distinct "connected components" (emotion segments) $n_c$ along the sequence $\tau$ decreases. Given the finite length $n$ of the sequence, there must be a critical threshold $\delta_d$ at which all predictions merge into one single emotional segment; thus, for every $\Delta d \geq \delta_d$, we have exactly one emotional state. This transition captures the boundary between local emotional fluctuations and a stable global emotion state.

Formally, $n_c$ can be viewed as a function of $\Delta d$, and the relationship satisfies:
\[
    \lim_{\Delta d \to 0^+} n_c = n, \quad\text{and}\quad \lim_{\Delta d \to +\infty} n_c = 1,
\]
as illustrated in Figure~\ref{fig:clustermetric}.  More clearly, Given predictions on consecutive segments $\tau = \{y_1, y_2, \dots, y_n\}$, define the number of connected components as a function of the threshold $\Delta d$ explicitly by:
\[
n_c(\Delta d) = 
\begin{cases}
n, & 0 < \Delta d < \min\limits_{1 \leq i < n} |y_{i+1}-y_i| \\[8pt]
k, & \min\limits_{\substack{1 \leq i < n}} |y_{i+1}-y_i| \leq \Delta d < \delta_d, \quad 1 < k < n \\[8pt]
1, & \Delta d \geq \delta_d
\end{cases}
\] Here, the critical threshold $\delta_d$ is explicitly defined as:
\[
\delta_d = \max_{1 \leq i < n} |y_{i+1}-y_i|.
\]

The resulting curve $n_c(\Delta d)$ intuitively describes how quickly local predictions consolidate into a coherent global emotional label as the threshold increases. A rapid convergence toward 1 indicates consistent and stable predictions over consecutive EEG segments, aligning well with the global emotion label. Conversely, a slow convergence signals frequent fluctuations and instability in predictions.

For compact summarization, one can utilize either the critical threshold $\delta_d$ (indicating the smallest threshold required to unify predictions into one emotional state) or the area under the curve $n_c(\Delta d)$, denoted $A_c$. Intuitively, a smaller value of $\delta_d$ or $A_c$ corresponds to more stable, realistic emotional predictions with fewer temporal inconsistencies, hence reflecting less severe TsDLI effects. If two models share the same critical threshold $\delta_d$, the model with a smaller $A_c$ converges faster and thus demonstrates greater stability and consistency.



\section{Experiments}\label{sec:exp}




Two public datasets are employed in the subsequent experiments:the DREAMER dataset \cite{katsigiannis2017dreamer} and the DEAP dataset \cite{koelstra2011deap}.  DREAMER comprises electroencephalogram (EEG, 14 channels, 128 Hz) and electrocardiogram (ECG, 2 channels, 256 Hz) signals collected from 23 subjects while they were exposed to 18 distinct audio-visual stimuli. After each clip, participants self-reported their emotional states using valence, arousal, and dominance scales (1-5 Likert scale). The DEAP dataset includes EEG (32 channels, 512 Hz downsampled to 128 Hz) and peripheral physiological signals (e.g., GSR, EMG) from 32 participants as they watched 40 one-minute music videos. Participants rated each video on valence, arousal, dominance, liking, and familiarity (1-9 scale). In our experiments, we focus solely on the EEG signals and dimensional arousal and valence labels.



\textbf{\textit{Experimental Settings}: }
There are typically two paradigms in emotion recognition: subject-independent (SI) and subject-dependent (SD) tasks. Although SI models are generally preferred for predictive applications, SD approaches provide a more controlled environment for analysis and are better suited for investigating the TsDLI problem addressed in this work. In SI tasks, different subjects often exhibit divergent self-assessments even when exposed to identical stimuli. This variability reflects the heterogeneous perception of emotional intensities among individuals and implies that the corresponding emotion graphs, as modeled in Section \ref{sec:method}, differ across subjects. Addressing these differences would require additional modeling, alignment, and evaluation efforts, thereby detracting from the primary focus of this paper.

In contrast, SD settings reduce inter-subject variability, allowing for a clearer focus on the TsDLI issues, particularly in scenarios where local emotion labels are unavailable. Moreover, the methods introduced in Section \ref{sec:related} under the LNL setting are more straightforward to implement in SD tasks than in SI tasks. For these reasons, the experiments reported in this section are conducted within a subject-dependent framework.

Following the subject-dependent paradigm, a separate model is trained for each subject---a strategy widely adopted in previous studies due to the non-stationarity of EEG signals and the variability of emotion label distributions across subjects. To ensure the reliability of our results, we perform 10 runs for each experimental setting, with different label noise introduced in each run. The reported outcomes represent the averaged results over these 10 runs and across all 23 subjects.

\textbf{\textit{Data Preprocessing:}} For each subject, the following procedure is applied to each trial's stimulus signals, denoted by $X_s$, and their corresponding baseline signals, $X_b$. In this context, the \emph{stimulus signals} (denoted by $X_s$) refer to the physiological recordings (e.g., EEG, ECG) captured while the subject is exposed to the experimental stimuli. In contrast, the \emph{baseline signals} (denoted by $X_b$) are recorded during a resting or neutral state—typically before the stimulus is presented—and are used to characterize the subject's normal activity. 

We use these to perform z-score normalization using the associated baseline signal:
\[
X_s' = \frac{X_s - \mu_b}{\sigma_b},
\]
where $\mu_b$ and $\sigma_b$ represent the mean and standard deviation of $X_b$, respectively. Next, we scale $X_s'$ to the range $[-1,1]$ by dividing it by its maximum absolute value:
\[
X_s'' = \frac{X_s'}{\max |X_s'|}.
\]
Subsequently, the processed signal $X_s''$ is partitioned into two non-overlapping temporal segments: a training set comprising the first 40 seconds and a test set comprising the last 20 seconds. Finally, each subset is segmented into 1-second non-overlapping windows with a stride of 1 second. This procedure is repeated for all 18 trials, resulting in training and test sets that collectively include data from every trial.

For each short segment, we assume it inherits the emotion label of its parent trial, as is customary in previous studies. In addition, we introduce label noise into the training set by randomly flipping a portion of the true labels to alternative labels with equal probability. In this work, we consider two noise levels: 20\% and 40\%.



\textbf{\textit{Backbones:}} With rapid advancements in deep learning, a wide array of neural network architectures has emerged, ranging from LSTM networks to transformer-based models. To demonstrate the generalizability of our proposed method across different architectures, we evaluate it on three representative models: EEGNet (a CNN-based model) \cite{lawhern2018eegnet}, an LSTM-based model, and a transformer-based model.


\textbf{\textit{Baselines:}} We compare our approach against several state-of-the-art methods for label-noise learning, as described in Section~\ref{sec:related}. Specifically, we evaluate five baseline methods: Regular (i.e., without any specialized label-noise handling), Co-teaching~\cite{han2018co}, SELF~\cite{nguyen2019self}, DivideMix~\cite{li2020dividemix}, CTW~\cite{ma2023ctw}, and StudentLoss~\cite{zhang2024student}.


\textbf{\textit{Evaluation Metrics:}} For a comprehensive assessment, we employ five distinct metrics: the F1 score, the Top-2 accuracy (Top2) score, the local fluctuation measure 
\[v_d := \frac{1}{N}\sum_{i=1}^N |y_{i+1}-y_{i}|,\]
and the metrics $A_c$ and $\delta_d$ introduced in Section~\ref{sec:eval}. The metric $v_d$ quantifies the cumulative magnitude of jumps between consecutive predictions, thereby directly capturing local fluctuations. The F1 score and Top-2 accuracy are focused on more quantitative assessments, while the rest of the metrics are more on the qualitative side. These metrics are subsequently used to rank the performance of different methods in our benchmark. For a detailed, comprehensive yet  still conclusive comparison, we adopt the weighted \textit{Borda Count} (B.C.) method \cite{saari1985optimal,alvo2014statistical,drotar2017heterogeneous} to aggregate the results from our extensive experiments. 

Assuming the $i$th metric $\mathcal{M}_i(Tr,S,Bn,Ec)$ whose value on test depends on Training method ($Tr$), actual subject ($S$), backbone network selected ($Bn$), Emotion category ($Ec$, valence or arousal), the mean over all subjects $\bar{\mathcal{M}}_i(Tr,Bn,Ec) = \sum_{S}\mathcal{M}_i(Tr,S,Bn,Ec)$ is obtained. Borda Count (B.C.) (with equal weights) is then performed to aggregate the influence on categorical variable $Bn$ and $Ec$ to get the rank $R_i(Tr)$ associated with the $i$th metric depending on $Tr$. The discrete distribution of ranks $\{R_i(Tr)\}_{i\in I}, I = \{F_1, top2, A_c, \delta_d, v_d\}$ can help reveal training method $Tr$'s performance span over these 5 metrics and gauge the balance between quantitative v.s qualitative metrics. 

Furthermore, these conditioned ranks can be further aggregated over different metrics to help provide a single overall rank. 
imbalanced nature of these metrics (2 are quantitative metrics and 3 are qualitative metrics), the two quantitative metrics (F1 score, Top-2 accuracy) are assigned a weight of 0.25 each, while three qualitative metrics ($A_c$, $v_d$, $\delta_d$) are weighted by $\frac{1}{6}$, ensuring equal total weights (0.5) for both the quantitative and qualitative evaluations---thus avoiding any inadvertent bias of one over the other. For example, if a method receives a ranking vector as $\{2,1,5,3,6\}$ for the considered metrics, the overall aggregated rank for it is then $2 * 0.25+ 1*0.25 + 5*\frac{1}{6} + 3*\frac{1}{6} + 6*\frac{1}{6} = \frac{37}{12} \approx 3.1$. This aggregated value will then be used for making the final rank table comparing with other methods. The resulting aggregated ranking offers a holistic evaluation by jointly considering both quantitative and qualitative aspects. Codes for the proposed methods and the evaluations will be made publicly available upon acceptance of this paper.

\begin{figure}[htbp]
    \centering
    \subfigure[Performance on 20\% symmetric noise (DREAMER).]{
        \includegraphics[width=0.45\textwidth]{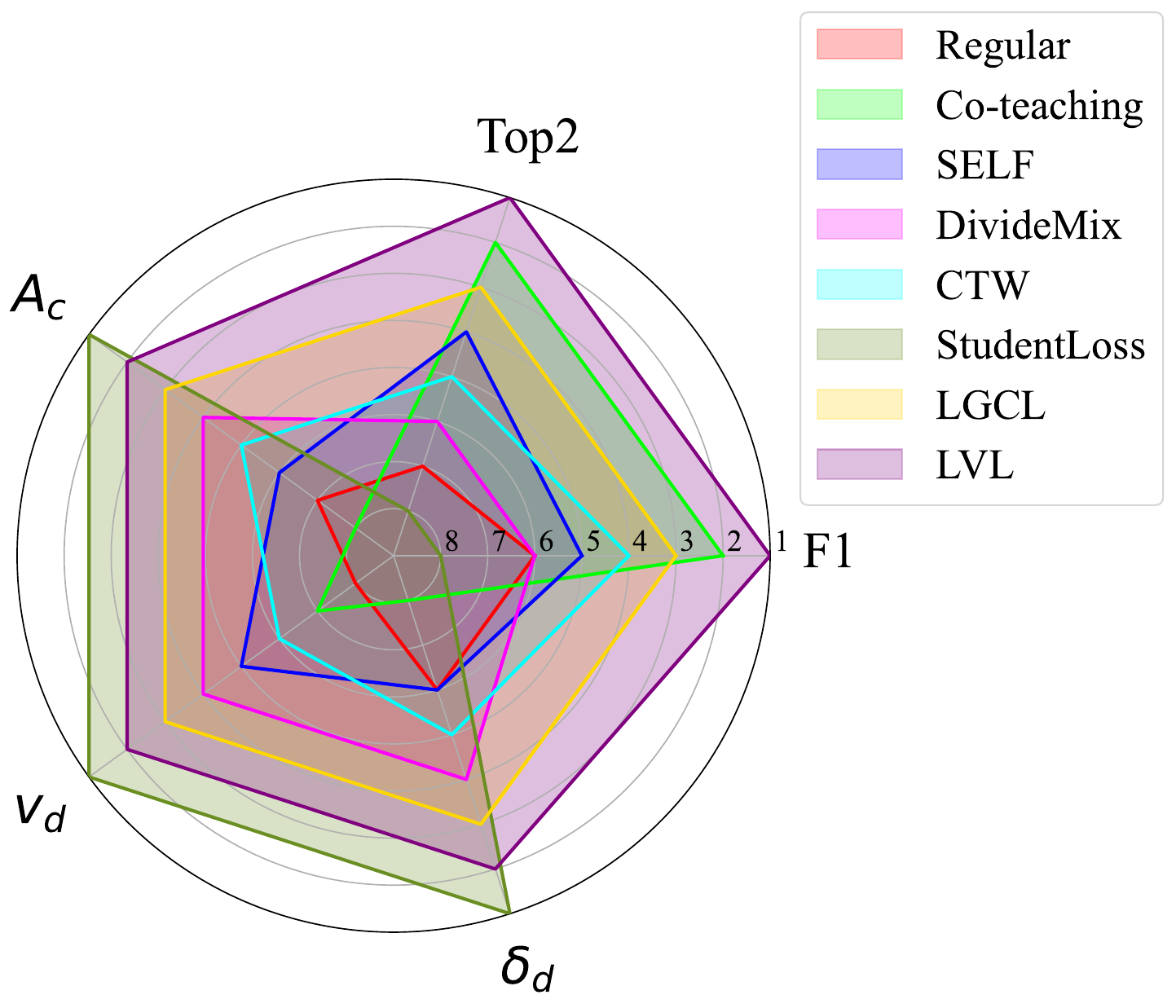} 
        \label{fig:image1}
    }
    \subfigure[Performance on 40\% symmetric noise (DREAMER).]{
        \includegraphics[width=0.45\textwidth]{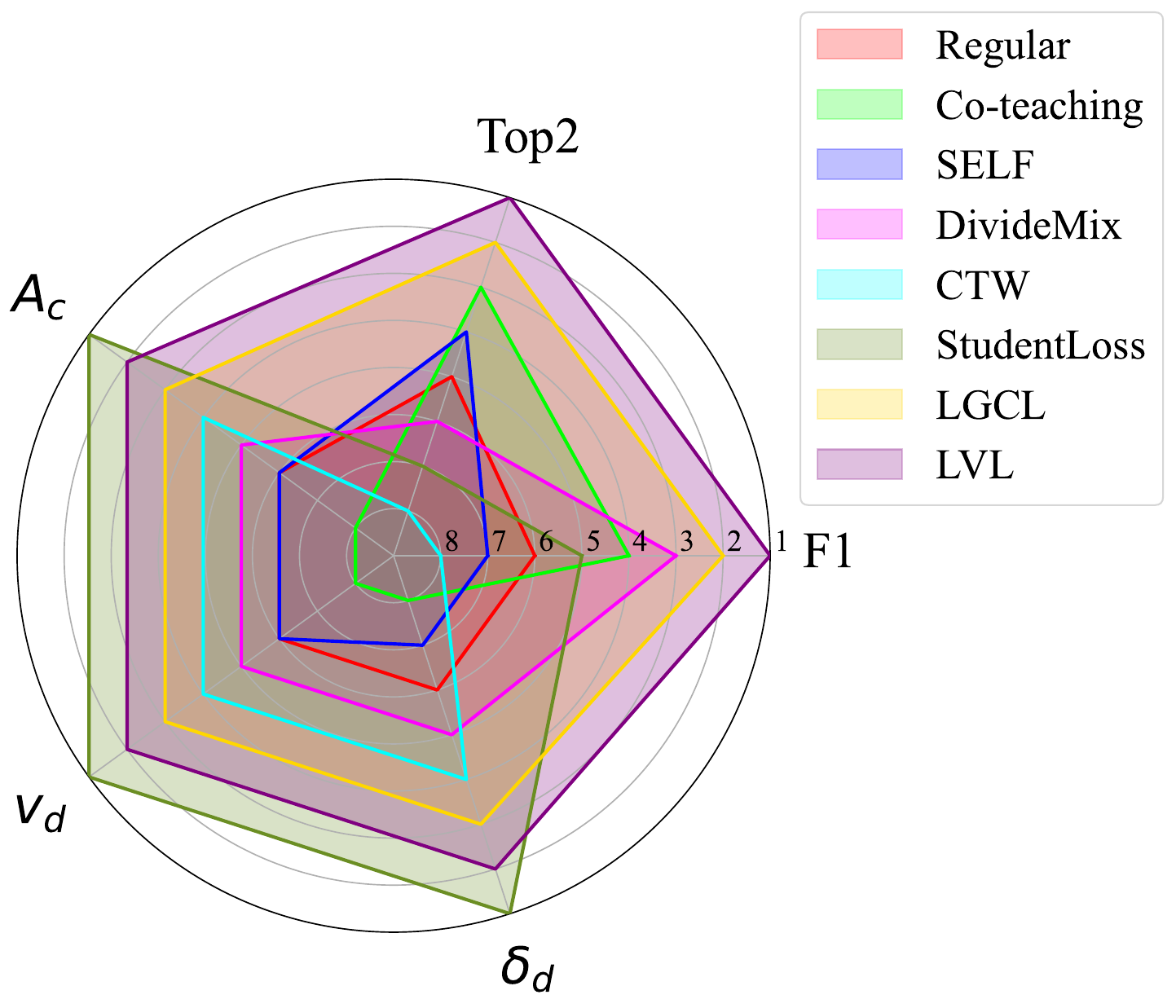} 
        \label{fig:image2}
    }
    \\
    \subfigure[Performance on 20\% symmetric noise (DEAP).]{
        \includegraphics[width=0.45\textwidth]{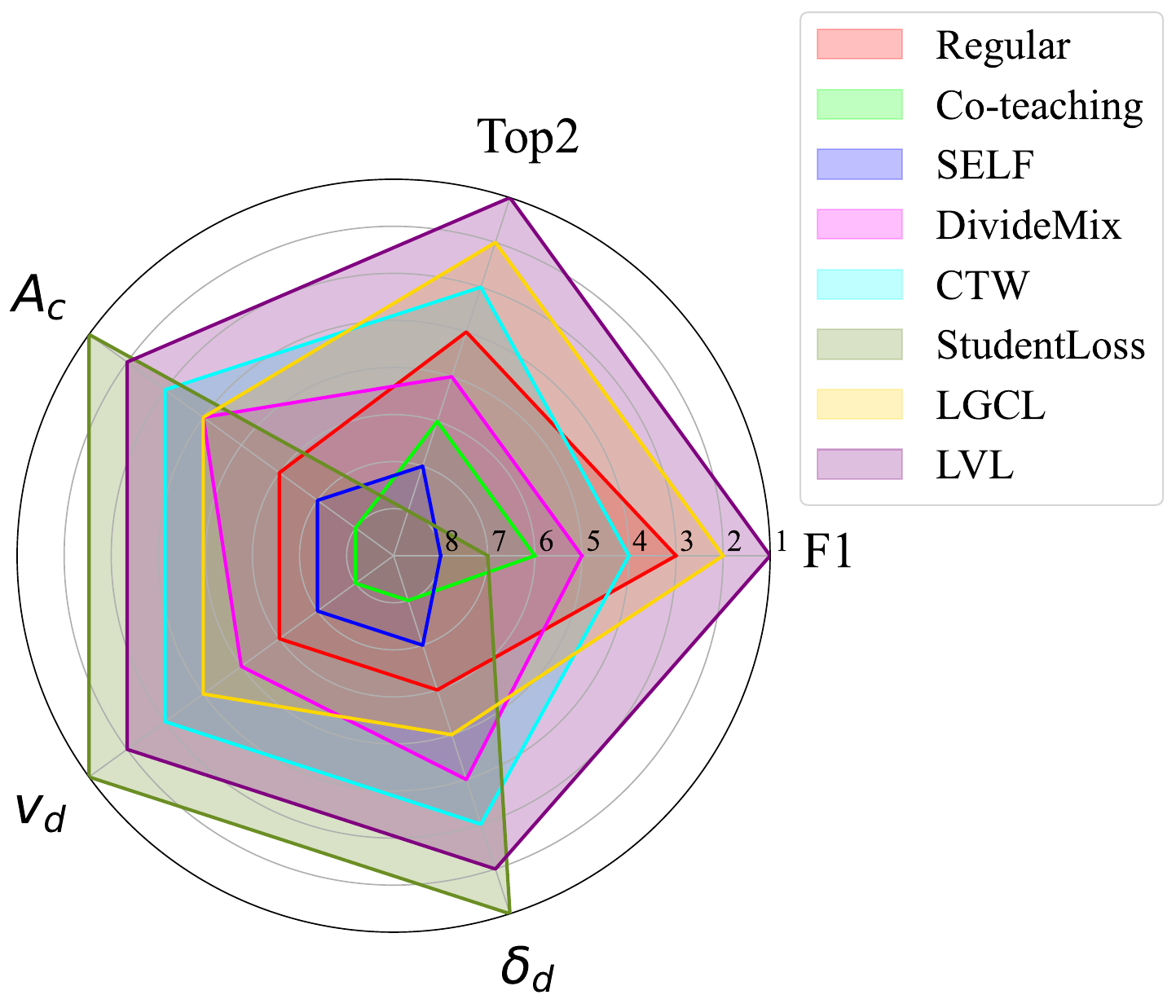} 
        \label{fig:image3}
    }
    \subfigure[Performance on 40\% symmetric noise (DEAP).]{
        \includegraphics[width=0.45\textwidth]{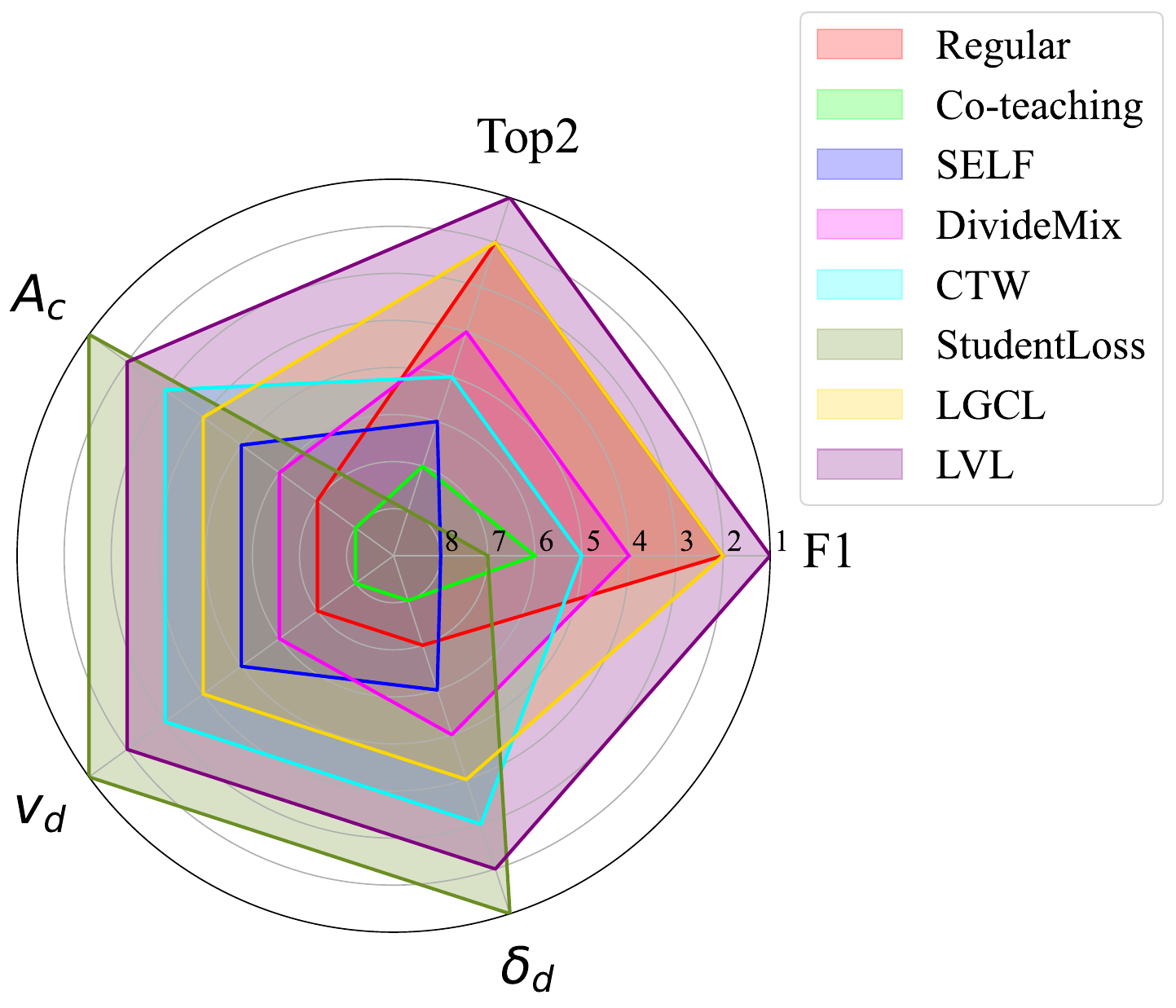} 
        \label{fig:image4}
    }
    \caption{Radar chart illustrating the performance of different methods across multiple evaluation metrics. Each axis represents a specific metric, and the radius denotes the relative ranking of each method (higher values indicate better ranks). The farther a method's polygon extends from the center, the better its overall performance across metrics.}
    \label{fig:both_images}
\end{figure}

\subsection{Main Results}\label{sec:main_result}
Figure \ref{fig:both_images} demonstrates the summarization of overall rank comparisons among different methods as radar plots ($R_i{Rr}$ will correspond to a polygon with given $Tr$ in each subplot. The detailed aggregated rank tables are listed in \ref{sec:rank_tables}). Considering the ranking nature of our evaluations and as a way to avoid possible label fluctuations causing non-informative conclusions (which usually happens in surveys adopting very fine scaled self-assessments as DEAP) \cite{saris2014design}, we apply a quartile-based relabeling of the original 10-point (0-9) self-reported scales. Specifically, labels are discretized into four bins (0-3) per subject, reflecting the quartile in which each score falls. This relabeling reduces noise from individual scale use variability and promotes more interpretable modeling of emotional responses grounded in neurophysiological signal variation.

Among all evaluated methods, our proposed models---especially the LVL variant---consistently achieve the most balanced performance across both quantitative and qualitative dimensions. This is evident from their broad and stable coverage across training regimes, noise levels, and both DREAMER and DEAP datasets. One notable exception occurs under 20\% noise in DEAP, where CTW performs comparably to LGCL, closely contending for second place. StudentLoss, by contrast, tends to favor qualitative coherence at the expense of quantitative fidelity, while Co-teaching shows the opposite trend---performing well quantitatively on DREAMER but failing to generalize robustly to DEAP. DivideMix, the most complex among the baselines, achieves moderately balanced results but still underperforms relative to our proposed regularization strategies. Other baseline methods demonstrate inconsistent behavior across evaluation metrics and training regimes, raising concerns about their reliability in real-world affective computing settings.

Overall, our methods offer the most stable and consistent (i.e. balanced) performance across both neurophysiologically relevant datasets and diverse modeling architectures. Their robustness across training regimes and resilience to label noise underscore their practical utility for EEG-based emotion recognition systems, where maintaining a balance between signal-driven accuracy and psychological interpretability is critical.

\subsection{An Ablation Study}\label{sec:ablation}

\textit{Effect of the graph configuration:} Table~\ref{tab:ablation1} compares how different prior emotion transition graphs affect the overall performance. Five graph configurations are evaluated: $G_0$ (the reference configuration), $G_1$ (in which edges $e_{1-2}$ and $e_{4-5}$ carry half the weight of $e_{2-3}$ and $e_{3-4}$), $G_2$ (where edges $e_{2-3}$ and $e_{3-4}$ carry half the weight of $e_{1-2}$ and $e_{4-5}$), $G_A$ (a complete graph with edges between all distinct vertices), and “w/o G” (no graph). Although no single configuration excels in all metrics, omitting the graph (w/o G) clearly degrades qualitative performance, indicating increased prediction fluctuations and label inconsistencies. Configuration $G_2$ produces the best qualitative metrics, which aligns with the “central tendency bias” described in Section~\ref{sec:PAET}. Specifically, reducing weights on transitions among mid-level states (2,3,4) lowers the cost of moderate state shifts, while transitions to extreme levels (1 or 5) remain relatively penalized, thus mitigating large prediction jumps.


\begin{table}[h]

    \centering

    \caption{Prediction performance when different prior graph configurations are adopted. Results are gathered when EEGNet is used as the backbone with 20\% noise in valence. The mean value and standard deviation across all subjects and all folds on the test set are shown. The best results per metric are highlighted in \textbf{\textcolor{green}{green}}, while the worst results are highlighted in \textbf{\textcolor{red}{red}}.}

    \label{tab:ablation1} 

    \begin{tabular}{c|l|cc|ccc}

        \toprule

        Method & Graph & F1 & Top2 & $A_c$ & $v_d$ & $\delta_d$ \\

        \midrule

        \multirow{5}{*}{LVL}

         & $G_0$ & 70.74 \(\pm\) 8.98 & \textbf{\textcolor{green}{87.52}} \(\pm\) 5.43 & 2.16 \(\pm\) 0.41 & 0.14 \(\pm\) 0.03 & 0.33 \(\pm\) 0.06 \\

         & $G_1$ & 70.99 \(\pm\) 9.05 & 87.28 \(\pm\) 5.42 & 2.10 \(\pm\) 0.41 & \textbf{\textcolor{green}{0.13}} \(\pm\) 0.03 & 0.32 \(\pm\) 0.06 \\

         & $G_2$ & \textbf{\textcolor{red}{70.49}} \(\pm\) 9.30 & \textbf{\textcolor{red}{86.78}} \(\pm\) 5.51 & \textbf{\textcolor{green}{2.03}} \(\pm\) 0.38 & \textbf{\textcolor{green}{0.13}} \(\pm\) 0.03 & \textbf{\textcolor{green}{0.31}} \(\pm\) 0.05 \\

         & $G_A$ & 71.40 \(\pm\) 8.92 & 87.44 \(\pm\) 5.31 & 2.10 \(\pm\) 0.50 & 0.13 \(\pm\) 0.03 & 0.32 \(\pm\) 0.07 \\

         & w/o $G$ & \textbf{\textcolor{green}{71.52}} \(\pm\) 8.84 & 87.31 \(\pm\) 5.22 & \textbf{\textcolor{red}{2.52}} \(\pm\) 0.41 & \textbf{\textcolor{red}{0.16}} \(\pm\) 0.03 & \textbf{\textcolor{red}{0.38}} \(\pm\) 0.06 \\

        \midrule

        \multirow{5}{*}{LGCL}

         & $G_0$ & 71.27 \(\pm\) 9.11 & \textbf{\textcolor{red}{87.05}} \(\pm\) 5.39 & 2.29 \(\pm\) 0.46 & 0.14 \(\pm\) 0.03 & 0.35 \(\pm\) 0.06 \\

         & $G_1$ & 71.32 \(\pm\) 8.95 & 87.37 \(\pm\) 5.47 & 2.20 \(\pm\) 0.41 & 0.14 \(\pm\) 0.03 & 0.33 \(\pm\) 0.06 \\

         & $G_2$ & \textbf{\textcolor{red}{71.14}} \(\pm\) 9.27 & 87.37 \(\pm\) 5.58 & \textbf{\textcolor{green}{2.13}} \(\pm\) 0.39 & \textbf{\textcolor{green}{0.13}} \(\pm\) 0.03 & \textbf{\textcolor{green}{0.32}} \(\pm\) 0.05 \\

         & $G_A$ & \textbf{\textcolor{green}{71.98}} \(\pm\) 8.92 & \textbf{\textcolor{green}{87.59}} \(\pm\) 5.35 & 2.23 \(\pm\) 0.50 & 0.14 \(\pm\) 0.03 & 0.33 \(\pm\) 0.07 \\

         & w/o $G$ & 71.44 \(\pm\) 8.55 & 87.45 \(\pm\) 5.29 & \textbf{\textcolor{red}{2.53}} \(\pm\) 0.42 & \textbf{\textcolor{red}{0.16}} \(\pm\) 0.03 & \textbf{\textcolor{red}{0.38}} \(\pm\) 0.06 \\

        \bottomrule

    \end{tabular}

\end{table} 

\textit{Combining LVL and LGCL} \quad Table \ref{tab:ablation2} collects the result where we combine both LVL and LGCL in the regularization as $\alpha$ LVL + $\beta$ LGCL, with different choices of the $\alpha$ and $\beta$ values. No statistically significant differences are observed, which is additional numerical support of the equivalency shown in the proof in Appendix \ref{appendix}.

\begin{table}[h]

    \centering

    \caption{Prediction performance when different combinations of weights are set between LVL and LGCL. Results gathered are shown when EEGNet is as the backbone with 20\% noise in valence. The best results per metric are in bold.}

    \label{tab:ablation2}

    \begin{tabular}{c|cc|ccc}

        \toprule

        $\alpha,\beta$ & F1 & Top2 & $A_c$ & $v_d$ & $\delta_d$\\

        \midrule  

        0.0, 1.0  & 71.27 $\pm$ 9.11 & \textbf{87.52} $\pm$ 5.39 & 2.29 $\pm$ 0.46 & 0.14 $\pm$ 0.03 & 0.35 $\pm$ 0.06 \\

        \midrule

        0.1, 0.9 & 71.08 $\pm$ 8.83 & 87.40 $\pm$ 5.10 & 2.18 $\pm$ 0.47 & 0.14 $\pm$ 0.03 & \textbf{0.33} $\pm$ 0.07 \\

        1/3, 2/3 & 71.04 $\pm$ 9.06 & 87.46 $\pm$ 5.19 & 2.22 $\pm$ 0.48 & 0.14 $\pm$ 0.03 & 0.34 $\pm$ 0.06 \\

        0.5, 0.5  & 71.34 $\pm$ 9.00 & 87.34 $\pm$ 5.20 & 2.23 $\pm$ 0.47 & 0.14 $\pm$ 0.03 & 0.34 $\pm$ 0.06 \\

        2/3, 1/3 & \textbf{71.64} $\pm$ 8.91 & 87.18 $\pm$ 5.14 & 2.26 $\pm$ 0.48 & 0.14 $\pm$ 0.03 & 0.34 $\pm$ 0.06 \\

        0.9, 0.1 & 71.16 $\pm$ 8.91 & 87.10 $\pm$ 5.54 & 2.29 $\pm$ 0.46 & 0.14 $\pm$ 0.03 & 0.35 $\pm$ 0.06 \\

        \midrule

        1.0, 0.0 & 70.74 $\pm$ 8.98 & 87.05 $\pm$ 5.43 & \textbf{2.16} $\pm$ 0.41 & 0.14 $\pm$ 0.03 & \textbf{0.33} $\pm$ 0.06 \\       

        \bottomrule

    \end{tabular}

\end{table}


\subsection{Case Study}\label{sec:case}

\begin{figure}
    \centering
    \includegraphics[width=0.85\textwidth]{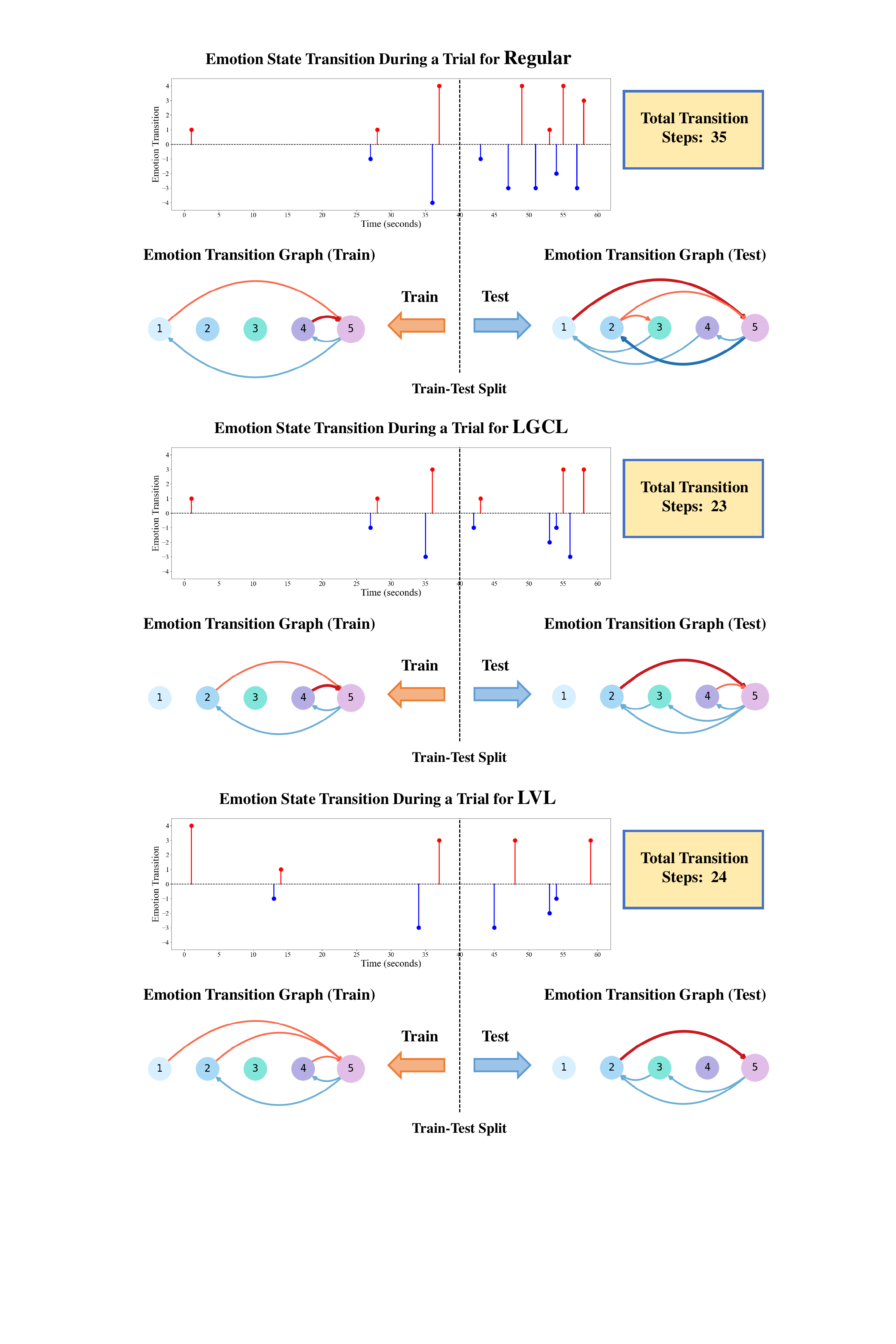}
    \caption{Case Study on emotion transitions for Subject~2's third trial in the first fold, as predicted by the LSTM-based model. In this trial, the global label is 5. In the directed graph, a thicker edge $i \to j$ indicates a higher frequency of the corresponding emotion transition in the prediction. The proposed LGCL and LVL regularizations effectively mitigate TsDLI issues.}
    \label{fig:case_study}
\end{figure}

To provide a direct visualization of how different training methods compare on the proposed qualitative metrics for the TsDLI issue, Figure~\ref{fig:case_study} presents an example from Subject~2's third trial using an LSTM-based backbone. In this example, we track the prediction difference between adjacent 1-second EEG segments, i.e., $y_{i+1} - y_i$, which is shown using stem plots. Additionally, the actual emotion state transitions are depicted as directed graphs below the stem plots. The visualization demonstrates that our regularization techniques effectively control both the number of fluctuations (total transition steps) and their amplitude, especially on the test set.


\begin{figure}[tbh]
    \centering
    \subfigure[ ]{
        \includegraphics[width=0.467\textwidth]{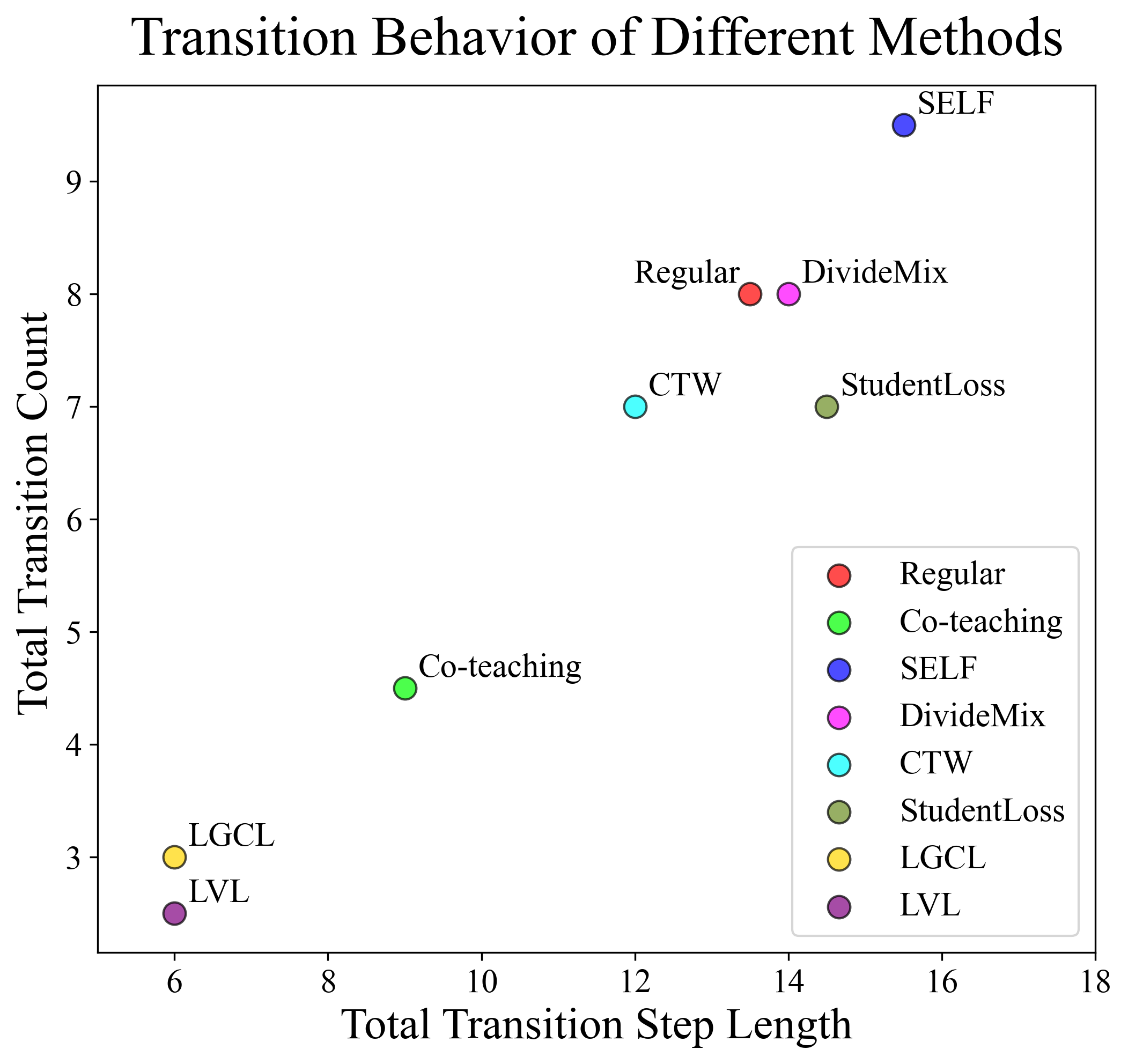}
        \label{fig:tt1}
    }
    \subfigure[ ]{
        \includegraphics[width=0.49\textwidth]{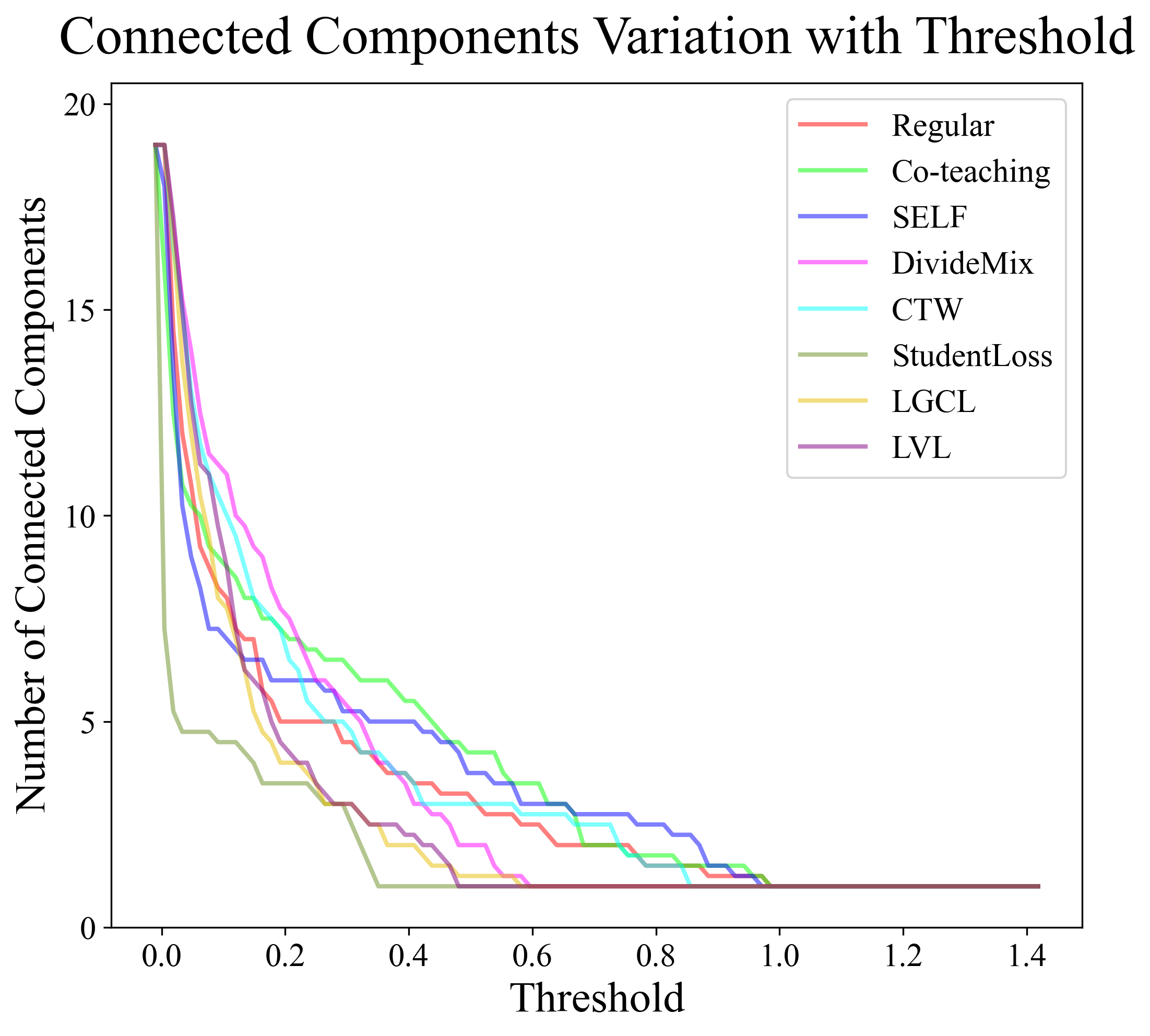}
        \label{fig:tt2}
    }
    \subfigure[ ]{
        \includegraphics[width=0.467\textwidth]{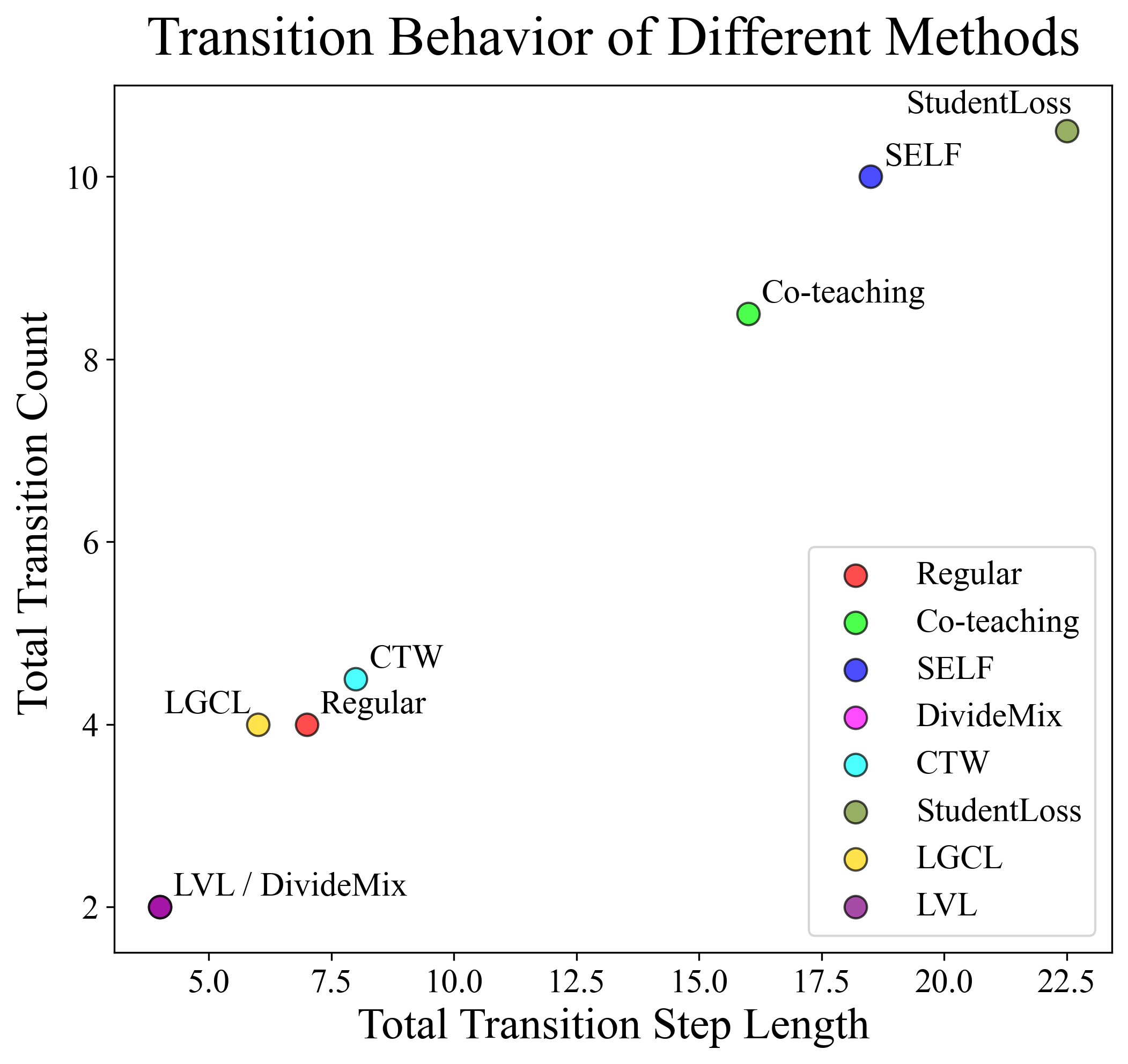}
        \label{fig:tt3}
    }
    \subfigure[ ]{
        \includegraphics[width=0.49\textwidth]{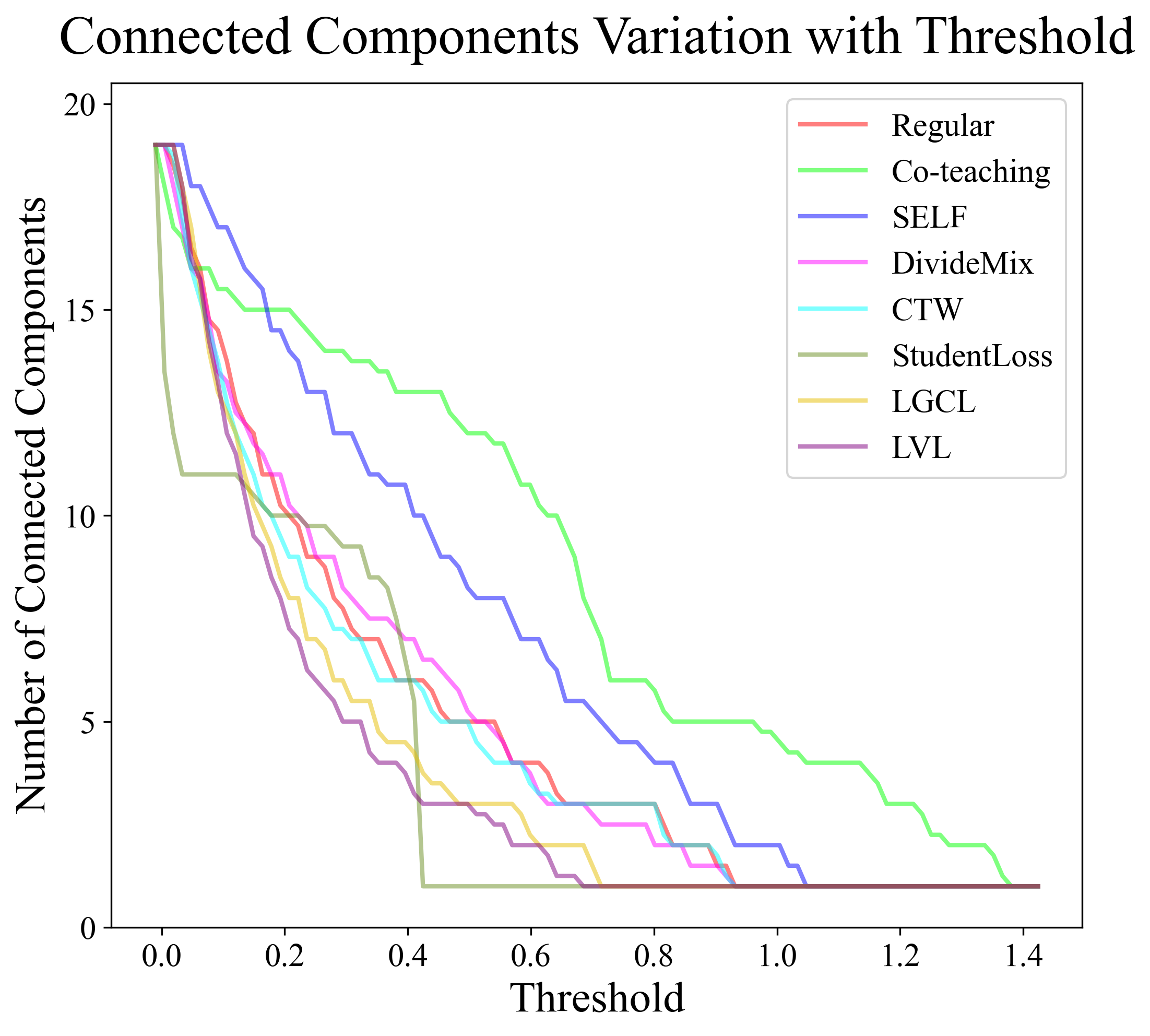}
        \label{fig:tt4}
    }
    
    \caption{Visualizations of transition behavior from differently trained models. On the left, the plot illustrates the total number of emotion transitions against the overall transition distance. On the right, the $n_c$ versus $\delta_d$ curves from various training strategies are displayed. Our proposed methods, LVL and LGCL, demonstrate superior performance on both metrics. These results are aggregated over all trials for Subject~1 in the first fold on valence, as predicted by the LSTM model, with median values reported. Plots on the top row are for DREAMER and the bottom plots are for DEAP.
}
    \label{fig:transition_and_connect}
\end{figure}

As another case study, Figure~\ref{fig:transition_and_connect} shows the relationship between the total transition count $\sum \delta(y_{i+1} \neq y_i)$ and the total transition length $\sum_{y_{i+1} \neq y_i} |y_{i+1} - y_i|$ (i.e., the sum of level changes for each transition), along with the corresponding $n_C$ versus $\Delta_d$ curves for various state-of-the-art methods. These visualizations clearly demonstrate that our metrics effectively capture local emotion transitions, and that our regularization techniques based on LVL and LGCL losses yield models with improved qualitative behavior in addressing the TsDLI issue. Notably, StudentLoss shows more transitions despite a faster decline in $n_C$, likely because it introduces a prior over the sample distribution, which induces underfitting and leads to less confident predictions near decision boundaries—making the model sensitive to small fluctuations.


\section{Discussion}\label{sec:discussion}





The foregoing experiments show that the proposed \emph{Local-Variation Loss} (LVL) and \emph{Local-Global Consistency Loss} (LGCL) offer a principled remedy to the TsDLI problem.  By embedding a clinically informed emotion-transition graph into the loss kernel, both terms regularize short-horizon predictions without requiring segment-level labels.  Across three backbones and two noise regimes (Figs.~\ref{fig:summary20}-\ref{fig:summary40}), LVL and LGCL jointly improve qualitative stability metrics ($A_c$, $v_d$, $\delta_d$) while preserving, and at times enhancing, quantitative scores (F1, Top-2).  The graph ablation in Table~\ref{tab:ablation1} further confirms that even coarse prior structure suppresses implausible prediction jitter.

This suppression of prediction jitter is handled differently by the different models.  That is, penalising commute distances between adjacent segments (LVL) implicitly models biological lags, while LGCL anchors micro-fluctuations to the macro-level appraisal.  The resulting trajectories (Fig.~\ref{fig:case_study}) display fewer improbable ``whiplash'' transitions that more appropriately due to the established notion of \emph{emotional inertia} \cite{kuppens2017emotion}.  Moreover, the dual-metric scheme exposes the trade-off between numerical accuracy and clinical credibility: methods that emphasize F1 alone (e.g. Co-teaching) may yield high accuracy yet unstable temporal paths, whereas overly conservative losses (e.g. StudentLoss) underfit and inflate transition counts.  LVL and LGCL provide what appear to be a \emph{pareto-optimal} \cite{pallasdies2021neural} compromise.

The proposed regularization strategy produces models with improved label consistency, enabling existing EEG datasets to be leveraged more effectively when training future foundation models for affective computing.  The same losses can also be applied during downstream adaptation of pretrained foundation models (e.g., \cite{yue2024eegpt})to tailor them to domain-specific tasks.  Beyond performance gains, the framework offers a principled tool for probing the temporal dynamics of human emotion, thereby supporting the design of affective systems that are more responsive, adaptive, and clinically consistent, ultimately enhancing user experience and well-being.

While the apporach in this work offers a plausible solution to the TsDLI problem, some of the limitations of the method include the following: (1) \emph{regularization versus local decoding}---LVL and LGCL temper learning dynamics but do \emph{not} explicitly infer time-resolved emotion labels; applications that demand explicit local affect estimates (e.g. biofeedback) will require an additional latent-state module (e.g. sequential VAEs, HMMs, etc.);
(2)	\emph{fixed graph priors}---all experiments in this work employ a hand-crafted line graph (Sec.~\ref{sec:PAET}), and as indicated in  Table~\ref{tab:ablation1} the graph topology is not unique; numerical performance and clinical plausibility must be balanced jointly, and therefore adaptive or data-driven graph learning could be a target for future work; (3) \emph{subject-dependent scope}---this study isolates TsDLI within the subject-dependent (SD) paradigm.  Extending to subject-independent (SI) settings will require data harmonization across heterogeneous transition graphs and accounting for inter-participant variability in reporting scales.  Simiarly, trial-independent or even dataset-independent tasks could be evaluated, further augmenting the space of supported scenarios; and (4) \emph{sparse ground truth}---publicly available datasets rarely contain dense, time-aligned emotion labels, limiting direct validation of local predictions and obscuring phenomena such as \emph{emotional inertia} (or the persistence of affect due to neural and autonomic lags).

The implications of addressing these limitations is largely dependent on adjustments, and/or enhancements to future data collection practices, where we suggest that the following data could be invaluable for training deeper, higher resolution, emotion recognition regimes:
\begin{enumerate}[i]
    \item \textbf{Include multi-axis, multi-rate annotation:} Supplement valence-arousal with dominance, tension and engagement, sampled at staggered cadences (e.g.\ 1~Hz for valence/arousal, 0.2~Hz for dominance) to capture disparate latencies.
    \item \textbf{Provide relative judgements:}  Combine sparse absolute Likert scores with denser pairwise comparisons to encode local affective contrast more efficiently \cite{song2018efficient,burkner2022information}, e.g., ask ``Which of these two (overlapping) snippets felt \emph{more positive}?''
    \item \textbf{Event-locked probes:}  Insert micro-surveys immediately after salient stimulus events to map onset lags and recovery times, directly parameterizing the subjects \emph{emotional inertia}.
    \item \textbf{Syncopated physiological sampling:} Align EEG with peripheral indices (e.g. skin conductance, HRV) at higher frequency during hypothesized transition windows to constrain future cross-modal graph models.
\end{enumerate}


\section{Conclusion}\label{sec:conclusion}

We introduce the Timescale-Dependent Label Inconsistency (TsDLI) problem---an important yet largely overlooked challenge in EEG-based emotion recognition.  
To tackle TsDLI, we devise two complementary regularization losses, grounded in bounded-variation theory and graph commute-time distances, and prove that they are equivalent under certain conditions.  Coupled with a new set of evaluation metrics, these losses provide a principled way to reconcile short-segment predictions with global self-reports.  
Experiments on the DREAMER and DEAP datasets, spanning several neural architectures, show that our method consistently surpasses state-of-the-art baselines.

Beyond this proof of concept, the framework lays a foundation for models that are more logically consistent, stable, and aligned with domain knowledge. Future work could incorporate adaptive graph learning to capture subject-specific or time-varying transition patterns, and apply causal-inference techniques to clarify how EEG dynamics drive emotional states---thereby enhancing both interpretability and robustness.  
Such advances promise to increase the reliability of emotion recognition systems in real-world settings.

More broadly, our results demonstrate that TsDLI can be mitigated by jointly optimizing \emph{quantitative accuracy} and \emph{qualitative plausibility}.  
The proposed \emph{Local-Variation Loss} (LVL) and \emph{Local–Global Consistency Loss} (LGCL) strike one effective balance.  
Promising extensions include hierarchical graphs that link micro-scale EEG rhythms with macro-context signals, Bayesian approaches that treat graph structure as a latent variable, and transfer-learning strategies that adapt graph parameters across users.  By explicitly accounting for timescale, inertia, and relative appraisal, these regularizers move EEG-based affective computing toward systems that are both \emph{trustworthy} and \emph{precise}.

\paragraph{Competing Interests}  
The authors declare no competing interests. 

\paragraph{Funding Statement}  
This research was supported in part by the Young Scientists Fund of the National Natural Science Foundation of China (NSFC) under grant No.12301677, National Natural Science Foundation of China under grant 12371103 and the Guangdong Natural Science Foundation under Grant 2025A1515010276.


\bibliographystyle{abbrvnat}
\bibliography{refs}  

\begin{thebibliography}{53}
\providecommand{\natexlab}[1]{#1}
\providecommand{\url}[1]{\texttt{#1}}
\expandafter\ifx\csname urlstyle\endcsname\relax
  \providecommand{\doi}[1]{doi: #1}\else
  \providecommand{\doi}{doi: \begingroup \urlstyle{rm}\Url}\fi

\bibitem[Alvo and Philip(2014)]{alvo2014statistical}
M.~Alvo and L.~Philip.
\newblock \emph{Statistical methods for ranking data}, volume 1341.
\newblock Springer, 2014.

\bibitem[Berthelot et~al.(2019)Berthelot, Carlini, Goodfellow, Papernot, Oliver, and Raffel]{berthelot2019mixmatch}
D.~Berthelot, N.~Carlini, I.~Goodfellow, N.~Papernot, A.~Oliver, and C.~A. Raffel.
\newblock Mixmatch: A holistic approach to semi-supervised learning.
\newblock \emph{Advances in neural information processing systems}, 32, 2019.

\bibitem[Buda et~al.(2018)Buda, Maki, and Mazurowski]{buda2018systematic}
T.~Buda, A.~Maki, and M.~A. Mazurowski.
\newblock A systematic study of the class imbalance problem in convolutional neural networks.
\newblock \emph{Neural Networks}, 106:\penalty0 249--259, 2018.

\bibitem[B{\"u}rkner(2022)]{burkner2022information}
P.-C. B{\"u}rkner.
\newblock On the information obtainable from comparative judgments.
\newblock \emph{psychometrika}, 87\penalty0 (4):\penalty0 1439--1472, 2022.

\bibitem[Cubuk et~al.(2019)Cubuk, Zoph, Mane, Vasudevan, and Le]{cubuk2019autoaugment}
E.~D. Cubuk, B.~Zoph, D.~Mane, V.~Vasudevan, and Q.~V. Le.
\newblock Autoaugment: Learning augmentation strategies from data.
\newblock In \emph{Proceedings of the IEEE Conference on Computer Vision and Pattern Recognition (CVPR)}, pages 113--123, 2019.

\bibitem[Cui et~al.(2020)Cui, Liu, Zhang, Chen, Wang, and Chen]{cui2020eeg}
H.~Cui, A.~Liu, X.~Zhang, X.~Chen, K.~Wang, and X.~Chen.
\newblock Eeg-based emotion recognition using an end-to-end regional-asymmetric convolutional neural network.
\newblock \emph{Knowledge-Based Systems}, 205:\penalty0 106243, 2020.

\bibitem[Davidson(1998)]{davidson1998affective}
R.~J. Davidson.
\newblock Affective style and affective disorders: Perspectives from affective neuroscience.
\newblock \emph{Cognition \& emotion}, 12\penalty0 (3):\penalty0 307--330, 1998.

\bibitem[Douven(2018)]{douven2018bayesian}
I.~Douven.
\newblock A bayesian perspective on likert scales and central tendency.
\newblock \emph{Psychonomic bulletin \& review}, 25:\penalty0 1203--1211, 2018.

\bibitem[Drot{\'a}r et~al.(2017)Drot{\'a}r, Gazda, and Gazda]{drotar2017heterogeneous}
P.~Drot{\'a}r, M.~Gazda, and J.~Gazda.
\newblock Heterogeneous ensemble feature selection based on weighted {B}orda {c}ount.
\newblock In \emph{2017 9th International Conference on Information Technology and Electrical Engineering (ICITEE)}, pages 1--4. IEEE, 2017.

\bibitem[Duan et~al.(2013)Duan, Zhu, and Lu]{duan2013differential}
R.-N. Duan, J.-Y. Zhu, and B.-L. Lu.
\newblock Differential entropy feature for {EEG}-based emotion classification.
\newblock In \emph{6th International IEEE/EMBS Conference on Neural Engineering (NER)}, pages 81--84. IEEE, 2013.

\bibitem[Furlanello et~al.(2018)Furlanello, Lipton, Tschannen, Itti, and Anandkumar]{furlanello2018born}
T.~Furlanello, Z.~C. Lipton, M.~Tschannen, L.~Itti, and A.~Anandkumar.
\newblock Born again neural networks.
\newblock In \emph{International Conference on Machine Learning (ICML)}, pages 1607--1616, 2018.

\bibitem[Ghosh et~al.(2017)Ghosh, Kumar, and Sastry]{ghosh2017robust}
A.~Ghosh, H.~Kumar, and P.~S. Sastry.
\newblock Robust loss functions under label noise for deep neural networks.
\newblock In \emph{Proceedings of the AAAI conference on artificial intelligence}, volume~31, 2017.

\bibitem[Goodfellow et~al.(2016)Goodfellow, Bengio, Courville, and Bengio]{goodfellow2016deep}
I.~Goodfellow, Y.~Bengio, A.~Courville, and Y.~Bengio.
\newblock \emph{Deep learning}, volume~1.
\newblock MIT press Cambridge, 2016.

\bibitem[Gutentag et~al.(2022)Gutentag, John, Gross, and Tamir]{gutentag2022incremental}
T.~Gutentag, O.~P. John, J.~J. Gross, and M.~Tamir.
\newblock Incremental theories of emotion across time: Temporal dynamics and correlates of change.
\newblock \emph{Emotion}, 22\penalty0 (6):\penalty0 1137, 2022.

\bibitem[Han et~al.(2018)Han, Yao, Yu, Niu, Xu, Hu, Tsang, and Sugiyama]{han2018co}
B.~Han, Q.~Yao, X.~Yu, G.~Niu, M.~Xu, W.~Hu, I.~Tsang, and M.~Sugiyama.
\newblock Co-teaching: Robust training of deep neural networks with extremely noisy labels.
\newblock \emph{Advances in neural information processing systems}, 31, 2018.

\bibitem[Janocha and Czarnecki(2017)]{janocha2017loss}
K.~Janocha and W.~M. Czarnecki.
\newblock On loss functions for deep neural networks in classification.
\newblock \emph{arXiv preprint arXiv:1702.05659}, 2017.

\bibitem[Katsigiannis and Ramzan(2017)]{katsigiannis2017dreamer}
S.~Katsigiannis and N.~Ramzan.
\newblock Dreamer: A database for emotion recognition through eeg and ecg signals from wireless low-cost off-the-shelf devices.
\newblock \emph{IEEE journal of biomedical and health informatics}, 22\penalty0 (1):\penalty0 98--107, 2017.

\bibitem[Klein and Randić(1993)]{Klein1993}
D.~J. Klein and M.~Randić.
\newblock Resistance distance.
\newblock \emph{Journal of Mathematical Chemistry}, 12:\penalty0 81--95, 1993.

\bibitem[Koelstra et~al.(2011)Koelstra, Muhl, Soleymani, Lee, Yazdani, Ebrahimi, Pun, Nijholt, and Patras]{koelstra2011deap}
S.~Koelstra, C.~Muhl, M.~Soleymani, J.-S. Lee, A.~Yazdani, T.~Ebrahimi, T.~Pun, A.~Nijholt, and I.~Patras.
\newblock Deap: A database for emotion analysis; using physiological signals.
\newblock \emph{IEEE transactions on affective computing}, 3\penalty0 (1):\penalty0 18--31, 2011.

\bibitem[Kuppens and Verduyn(2017)]{kuppens2017emotion}
P.~Kuppens and P.~Verduyn.
\newblock Emotion dynamics.
\newblock \emph{Current Opinion in Psychology}, 17:\penalty0 22--26, 2017.

\bibitem[Lawhern et~al.(2018)Lawhern, Solon, Waytowich, Gordon, Hung, and Lance]{lawhern2018eegnet}
V.~J. Lawhern, A.~J. Solon, N.~R. Waytowich, S.~M. Gordon, C.~P. Hung, and B.~J. Lance.
\newblock Eegnet: a compact convolutional neural network for eeg-based brain--computer interfaces.
\newblock \emph{Journal of neural engineering}, 15\penalty0 (5):\penalty0 056013, 2018.

\bibitem[Li et~al.(2021)Li, Xu, Wu, et~al.]{li2021learning}
F.~Li, H.~Xu, C.~Wu, et~al.
\newblock Learning from noisy labels with distillation.
\newblock In \emph{International Joint Conference on Artificial Intelligence (IJCAI)}, 2021.

\bibitem[Li et~al.(2020)Li, Socher, and Hoi]{li2020dividemix}
J.~Li, R.~Socher, and S.~C. Hoi.
\newblock Dividemix: Learning with noisy labels as semi-supervised learning.
\newblock \emph{arXiv preprint arXiv:2002.07394}, 2020.

\bibitem[Li et~al.(2022)Li, Zhang, Tiwari, Song, Hu, Yang, Zhao, Kumar, and Marttinen]{li2022eeg}
X.~Li, Y.~Zhang, P.~Tiwari, D.~Song, B.~Hu, M.~Yang, Z.~Zhao, N.~Kumar, and P.~Marttinen.
\newblock {EEG} based emotion recognition: A tutorial and review.
\newblock \emph{ACM Computing Surveys}, 55\penalty0 (4):\penalty0 1--57, 2022.

\bibitem[Liu et~al.(2024)Liu, Zhou, Zhu, Zhai, Jia, and Liu]{liu2024vbh}
C.~Liu, X.~Zhou, Z.~Zhu, L.~Zhai, Z.~Jia, and Y.~Liu.
\newblock Vbh-gnn: variational bayesian heterogeneous graph neural networks for cross-subject emotion recognition.
\newblock In \emph{The Twelfth International Conference on Learning Representations}, 2024.

\bibitem[Liu et~al.(2023)Liu, Chen, Pei, Ma, et~al.]{liu2023scale}
Z.~Liu, D.~Chen, W.~Pei, Q.~Ma, et~al.
\newblock Scale-teaching: Robust multi-scale training for time series classification with noisy labels.
\newblock \emph{Advances in Neural Information Processing Systems}, 36:\penalty0 33726--33757, 2023.

\bibitem[Lov{\'a}sz(1993)]{lovasz1993random}
L.~Lov{\'a}sz.
\newblock Random walks on graphs.
\newblock \emph{Combinatorics, Paul erdos is eighty}, 2\penalty0 (1-46):\penalty0 4, 1993.

\bibitem[Ma et~al.(2023)Ma, Liu, Zheng, Wang, and Ma]{ma2023ctw}
P.~Ma, Z.~Liu, J.~Zheng, L.~Wang, and Q.~Ma.
\newblock {CTW}: {C}onfident {T}ime-{W}arping for {T}ime-{S}eries {L}abel-{N}oise {L}earning.
\newblock In \emph{IJCAI}, pages 4046--4054, 2023.

\bibitem[Nguyen et~al.(2019)Nguyen, Mummadi, Ngo, Nguyen, Beggel, and Brox]{nguyen2019self}
D.~T. Nguyen, C.~K. Mummadi, T.~P.~N. Ngo, T.~H.~P. Nguyen, L.~Beggel, and T.~Brox.
\newblock Self: Learning to filter noisy labels with self-ensembling.
\newblock \emph{arXiv preprint arXiv:1910.01842}, 2019.

\bibitem[Pallasdies et~al.(2021)Pallasdies, Norton, Schleimer, and Schreiber]{pallasdies2021neural}
F.~Pallasdies, P.~Norton, J.-H. Schleimer, and S.~Schreiber.
\newblock Neural optimization: Understanding trade-offs with pareto theory.
\newblock \emph{Current opinion in neurobiology}, 71:\penalty0 84--91, 2021.

\bibitem[Patrini et~al.(2017)Patrini, Rozza, Menon, Nock, and Qu]{patrini2017making}
G.~Patrini, A.~Rozza, A.~K. Menon, R.~Nock, and L.~Qu.
\newblock Making deep neural networks robust to label noise: A loss correction approach.
\newblock In \emph{Proceedings of the IEEE Conference on Computer Vision and Pattern Recognition (CVPR)}, pages 1944--1952, 2017.

\bibitem[Paulhus et~al.(2007)Paulhus, Vazire, et~al.]{paulhus2007self}
D.~L. Paulhus, S.~Vazire, et~al.
\newblock The self-report method.
\newblock \emph{Handbook of research methods in personality psychology}, 1\penalty0 (2007):\penalty0 224--239, 2007.

\bibitem[Puccetti et~al.(2022)Puccetti, Villano, Fadok, and Heller]{puccetti2022temporal}
N.~A. Puccetti, W.~J. Villano, J.~P. Fadok, and A.~S. Heller.
\newblock Temporal dynamics of affect in the brain: Evidence from human imaging and animal models.
\newblock \emph{Neuroscience \& Biobehavioral Reviews}, 133:\penalty0 104491, 2022.

\bibitem[Ren et~al.(2018)Ren, Zeng, Yang, and Urtasun]{ren2018learning}
M.~Ren, W.~Zeng, B.~Yang, and R.~Urtasun.
\newblock Learning to reweight examples for robust deep learning.
\newblock In \emph{International Conference on Machine Learning (ICML)}, pages 4334--4343, 2018.

\bibitem[Saari(1985)]{saari1985optimal}
D.~G. Saari.
\newblock The optimal ranking method is the {B}orda {C}ount.
\newblock Technical report, Discussion paper, 1985.

\bibitem[Saris and Gallhofer(2014)]{saris2014design}
W.~E. Saris and I.~N. Gallhofer.
\newblock \emph{Design, evaluation, and analysis of questionnaires for survey research}.
\newblock John Wiley \& Sons, 2014.

\bibitem[Scherer(2009)]{scherer2009dynamic}
K.~R. Scherer.
\newblock The dynamic architecture of emotion: Evidence for the component process model.
\newblock \emph{Cognition and emotion}, 23\penalty0 (7):\penalty0 1307--1351, 2009.

\bibitem[Song et~al.(2021)Song, Choi, and Lee]{song2021multiclass}
H.~Song, J.~Choi, and I.~Y. Lee.
\newblock Multi-class learning from noisy partial labels.
\newblock \emph{IEEE Transactions on Neural Networks and Learning Systems}, 32\penalty0 (12):\penalty0 5600--5613, 2021.

\bibitem[Song et~al.(2022)Song, Kim, Park, Shin, and Lee]{song2022learning}
H.~Song, M.~Kim, D.~Park, Y.~Shin, and J.-G. Lee.
\newblock Learning from noisy labels with deep neural networks: A survey.
\newblock \emph{IEEE transactions on neural networks and learning systems}, 34\penalty0 (11):\penalty0 8135--8153, 2022.

\bibitem[Song et~al.(2018)Song, Li, Jia, Wang, and Rao]{song2018efficient}
R.~Song, Y.~Li, Y.~Jia, Y.~Wang, and P.~Rao.
\newblock Efficient, robust and divisible paired comparison for subjective quality assessment.
\newblock \emph{Multimedia Tools and Applications}, 77:\penalty0 13597--13613, 2018.

\bibitem[Tanaka et~al.(2018)Tanaka, Ikami, Yamasaki, and Aizawa]{tanaka2018joint}
D.~Tanaka, D.~Ikami, T.~Yamasaki, and K.~Aizawa.
\newblock Joint optimization framework for learning with noisy labels.
\newblock In \emph{Proceedings of the IEEE Conference on Computer Vision and Pattern Recognition (CVPR)}, pages 5552--5560, 2018.

\bibitem[Tarvainen and Valpola(2017)]{tarvainen2017mean}
A.~Tarvainen and H.~Valpola.
\newblock Mean teachers are better role models: Weight-averaged consistency targets improve semi-supervised deep learning results.
\newblock In \emph{Advances in Neural Information Processing Systems (NeurIPS)}, pages 1195--1204, 2017.

\bibitem[Tourangeau et~al.(2000)Tourangeau, Rips, and Rasinski]{tourangeau2000psychology}
R.~Tourangeau, L.~J. Rips, and K.~Rasinski.
\newblock \emph{The psychology of survey response}.
\newblock Cambridge University Press, 2000.

\bibitem[Von~Luxburg(2007)]{von2007tutorial}
U.~Von~Luxburg.
\newblock A tutorial on spectral clustering.
\newblock \emph{Statistics and computing}, 17:\penalty0 395--416, 2007.

\bibitem[Wang et~al.(2020)Wang, Ma, Zhao, and Tian]{wang2020comprehensive}
Q.~Wang, Y.~Ma, K.~Zhao, and Y.~Tian.
\newblock A comprehensive survey of loss functions in machine learning.
\newblock \emph{Annals of Data Science}, pages 1--26, 2020.

\bibitem[Wang et~al.(2019)Wang, Ma, Chen, Luo, Yi, and Bailey]{wang2019symmetric}
Y.~Wang, X.~Ma, Z.~Chen, Y.~Luo, J.~Yi, and J.~Bailey.
\newblock Symmetric cross entropy for robust learning with noisy labels.
\newblock In \emph{Proceedings of the IEEE/CVF international conference on computer vision}, pages 322--330, 2019.

\bibitem[Wang et~al.(2025)Wang, Sun, Zhou, Wang, Fan, Huang, and Bu]{wang2025noisygl}
Z.~Wang, D.~Sun, S.~Zhou, H.~Wang, J.~Fan, L.~Huang, and J.~Bu.
\newblock Noisy{GL}: {A} {C}omprehensive {B}enchmark for {G}raph {N}eural {N}etworks under {L}abel {N}oise.
\newblock \emph{Advances in Neural Information Processing Systems}, 37:\penalty0 38142--38170, 2025.

\bibitem[Waugh et~al.(2015)Waugh, Shing, and Avery]{waugh2015temporal}
C.~E. Waugh, E.~Z. Shing, and B.~M. Avery.
\newblock Temporal dynamics of emotional processing in the brain.
\newblock \emph{Emotion Review}, 7\penalty0 (4):\penalty0 323--329, 2015.

\bibitem[Xie et~al.(2020)Xie, Luong, Hovy, and Le]{xie2020self}
Q.~Xie, M.-T. Luong, E.~Hovy, and Q.~V. Le.
\newblock Self-training with noisy student improves imagenet classification.
\newblock In \emph{Proceedings of the IEEE/CVF Conference on Computer Vision and Pattern Recognition (CVPR)}, pages 10687--10698, 2020.

\bibitem[Yue et~al.(2024)Yue, Xue, Gao, Tang, Guo, Jiang, and Liu]{yue2024eegpt}
T.~Yue, S.~Xue, X.~Gao, Y.~Tang, L.~Guo, J.~Jiang, and J.~Liu.
\newblock Eegpt: Unleashing the potential of eeg generalist foundation model by autoregressive pre-training.
\newblock \emph{arXiv preprint arXiv:2410.19779}, 2024.

\bibitem[Zhang et~al.(2024)Zhang, Li, Fujita, Li, Wang, Zhu, Zhang, and Liu]{zhang2024student}
S.~Zhang, J.-Q. Li, H.~Fujita, Y.-W. Li, D.-B. Wang, T.-T. Zhu, M.-L. Zhang, and C.-Y. Liu.
\newblock Student loss: Towards the probability assumption in inaccurate supervision.
\newblock \emph{IEEE Transactions on Pattern Analysis and Machine Intelligence}, 46\penalty0 (6):\penalty0 4460--4475, 2024.

\bibitem[Zhang and Sabuncu(2018)]{zhang2018generalized}
Z.~Zhang and M.~Sabuncu.
\newblock Generalized cross entropy loss for training deep neural networks with noisy labels.
\newblock \emph{Advances in neural information processing systems}, 31, 2018.

\bibitem[Zheng and Lu(2015)]{zheng2015investigating}
W.-L. Zheng and B.-L. Lu.
\newblock Investigating critical frequency bands and channels for eeg-based emotion recognition with deep neural networks.
\newblock \emph{IEEE Transactions on autonomous mental development}, 7\penalty0 (3):\penalty0 162--175, 2015.

\end{thebibliography}

\appendix




\newpage

\section{Relevant Proofs}\label{appendix}

\begin{lemma}
\label{lemma 1}
If $L \in \mathbb{R}^{n \times n}$ is positive semi-definite, then the pseudo-inverse $L^{\dagger}$ is positive semi-definite.
\end{lemma}

\begin{proof}

Since $L$ is positive semi-definite, there exists an orthogonal matrix $U \in \mathbb{R}^{n \times n}$ s.t. 
\begin{align}
L = U^{T} \begin{bmatrix} \Sigma & 0 \\ 0 & 0 \end{bmatrix} U, 
\end{align}
where $\Sigma = \begin{bmatrix}
    d_1 & & &\\
    & d_2 & &\\
    & & \ddots &\\
    & & & d_r
\end{bmatrix}, \quad r \leq n, \; d_i > 0, \; i = 1, \cdots ,r.$

Then we prove that $L^{\dagger} = U^T 
\begin{bmatrix} \Sigma^{-1} & 0 \\ 0 & 0 \end{bmatrix} U.$

\begin{align}
&L L^\dagger L = U^{T} \begin{bmatrix} \Sigma & 0 \\ 0 & 0 \end{bmatrix} U U^{T} \begin{bmatrix} \Sigma^{-1} & 0 \\ 0 & 0 \end{bmatrix} U U^{T} \begin{bmatrix} \Sigma & 0 \\ 0 & 0 \end{bmatrix} U = U^{T} \begin{bmatrix} \Sigma & 0 \\ 0 & 0 \end{bmatrix} U = L
\\
&L^\dagger L L^\dagger = U^{T} \begin{bmatrix} \Sigma^{-1} & 0 \\ 0 & 0 \end{bmatrix} U U^{T} \begin{bmatrix} \Sigma & 0 \\ 0 & 0 \end{bmatrix} U U^{T} \begin{bmatrix} \Sigma^{-1} & 0 \\ 0 & 0 \end{bmatrix} U = U^{T} \begin{bmatrix} \Sigma^{-1} & 0 \\ 0 & 0 \end{bmatrix} U = L^\dagger
\\
&(L L^\dagger)^T = (U^{T} \begin{bmatrix} \Sigma^{-1} & 0 \\ 0 & 0 \end{bmatrix} U U^{T} \begin{bmatrix} \Sigma & 0 \\ 0 & 0 \end{bmatrix} U)^T = (U^{T} \begin{bmatrix} I & 0 \\ 0 & 0 \end{bmatrix} U)^T = U^{T} \begin{bmatrix} I & 0 \\ 0 & 0 \end{bmatrix} U = L L^\dagger
\\
&(L^\dagger L)^T = (U^{T} \begin{bmatrix} \Sigma & 0 \\ 0 & 0 \end{bmatrix} U U^{T} \begin{bmatrix} \Sigma^{-1} & 0 \\ 0 & 0 \end{bmatrix} U)^T = (U^{T} \begin{bmatrix} I & 0 \\ 0 & 0 \end{bmatrix} U)^T = U^{T} \begin{bmatrix} I & 0 \\ 0 & 0 \end{bmatrix} U = L^\dagger L
\end{align}


Since $d_i > 0$, $\Sigma^{-1}$'s diagonal elements $d_i^{-1} > 0$, therefore $L^\dagger$ is positive semi-definite.

\end{proof}

Before proving the equivalence result for the proposed two losses, we  need another classical result using the Cauchy-Schwarz inequality, which generically states that for any two vectors $x, y \in \mathbb{R}^n$:
\begin{equation}
    |x^T y|^2 \leq ||x||_2^2 ||y||_2^2
\end{equation}
In particular, if $y$ is a vector of all-one's, we have
\begin{equation}
    |\sum\limits_{i=1}^n x_i|^2 \leq n ||x||_2^2
\end{equation}

\begin{theorem}
    If $L$ is positive semi-definite, $L^\dagger$ is the pseudo-inverse of $L$, 
    
    \begin{align} 
    L_1 &= \frac{1}{N} \sum\limits_{i=2}^N \left( \hat{Y}(t_i) - \hat{Y}(t_{i-1}) 
 \right)^T  L^{{\dagger}} \left( \hat{Y}(t_i) - \hat{Y}(t_{i-1}) \right),
 \\ 
 L_2 &= \frac{1}{N} \sum\limits_{i=1}^N \left( \hat{Y}(t_i) - E[\hat{Y}(t)] \right)^T  L^{{\dagger}} \left(\hat{Y}(t_i) - E[\hat{Y}(t)] \right), 
 \end{align}

 Then $L_1$ and $L_2$ are equivalent in the sense that there exist two constants $C$ and $C^*$ (may depends on $N$) such that $L_1 \leq C L_2$ and $L_2 \leq C^* L_1$.
 
\end{theorem}





\begin{proof}
    By Lemma~\ref{lemma 1}, $L^\dagger$ is positive semi-definite. Therefore, there exists a matrix $P \in \mathbb{R}^{n \times n}$ such that $L = P^T P$.

    \begin{align}
    L_1^i &= \left( \hat{Y}(t_i) - \hat{Y}(t_{i-1}) \right)^T L^\dagger \left( \hat{Y}(t_i) - \hat{Y}(t_{i-1}) \right) \\
    &= \left( \hat{Y}(t_i) - \hat{Y}(t_{i-1}) \right)^T P^T P \left( \hat{Y}(t_i) - \hat{Y}(t_{i-1}) \right) \\
    &= \left( P \left( \hat{Y}(t_i) - \hat{Y}(t_{i-1}) \right) \right)^T \left( P \left( \hat{Y}(t_i) - \hat{Y}(t_{i-1}) \right) \right) \\
    &= \left\Vert P \left( \hat{Y}(t_i) - \hat{Y}(t_{i-1}) \right) \right\Vert_2^2\\
    L_2^i &= \left( \hat{Y}(t_i) - E[\hat{Y}(t)] \right)^T L^\dagger \left( \hat{Y}(t_i) - E[\hat{Y}(t)] \right) \\
    &= \left( \hat{Y}(t_i) - E[\hat{Y}(t)] \right)^T P^T P \left( \hat{Y}(t_i) - E[\hat{Y}(t)] \right) \\
    &= \left( P \left( \hat{Y}(t_i) - E[\hat{Y}(t)] \right) \right)^T \left( P \left( \hat{Y}(t_i) - E[\hat{Y}(t)]) \right) \right) \\
    &= \left\Vert P \left( \hat{Y}(t_i) - E[\hat{Y}(t)] \right) \right\Vert_2^2
    \end{align}

    Now, define $Z_i = P \hat{Y}(t_i)$, $E[Z] = E[P \hat{Y}(t)]$, then:
    \begin{align}
    L_1^i &= \left\Vert P \hat{Y}(t_i) - P \hat{Y}(t_{i-1}) \right\Vert_2^2 = \left\Vert Z_i - Z_{i-1} \right\Vert_2^2\\
    L_2^i &= \left\Vert P \hat{Y}(t_i) - P E[\hat{Y}(t)] \right\Vert_2^2 = \left\Vert P \hat{Y}(t_i) - E[P\hat{Y}(t)] \right\Vert_2^2\\ 
    &= \left\Vert Z_i - E[Z] \right\Vert_2^2 = \left\Vert Z_i - \frac{1}{N} \sum_{k=1}^{N} Z_k \right\Vert_2^2\\
    L_1^i &= \left\Vert Z_i - Z_{i-1} \right\Vert_2^2 = \left\Vert Z_i - E[Z] + E[Z] - Z_{i-1} \right\Vert_2^2 \\
    &\leq 2 \left \Vert Z_i - E[Z] \right\Vert_2^2 + 2 \left\Vert Z_{i-1} - E[Z] \right\Vert_2^2
\end{align}
We then can prove the first inequality:
\begin{align}
    L_1 &= \frac{1}{N} \sum_{i=2}^N L_1^i \leq \frac{2}{N} \sum_{i=2}^N 
  \left(\left\Vert Z_i - E[Z] \right\Vert_2^2 + \left\Vert Z_{i-1} - E[Z] \right\Vert_2^2\right) \\
    & = 2(L_2 - \frac{1}{N} \left\Vert Z_1 - E[Z] \right\Vert_2^2) + 2(L_2 - \frac{1}{N} \left\Vert Z_N - E[Z] \right\Vert_2^2) \leq 4 L_2
\end{align}

For the second inequality, we have:
\begin{align}
    L_2^i &= \left\Vert Z_i - E[Z] \right\Vert_2^2 = \left\Vert Z_i - \frac{1}{N} \sum_{k=1}^{N} Z_k \right\Vert_2^2 \\
    &= \frac{1}{N^2} \left\Vert \sum_{k=1}^{N} \left( Z_i - Z_k \right) \right\Vert_2^2 \leq \frac{1}{N} \sum_{k=1}^{N} \left\Vert Z_i - Z_k \right\Vert_2^2
\end{align}

Summing over all terms of $L_2^i$ gives:
\begin{align}
L_2 &= \frac{1}{N} \sum_{i=1}^N L_2^i \leq \frac{1}{N} \sum_{i=1}^N \frac{1}{N} \sum_{k=1}^{N} \left\Vert Z_i - Z_k \right\Vert_2^2\\
& = \frac{2}{N^2}\sum_{i<k}\left\Vert Z_i - Z_k \right\Vert_2^2
\leq \frac{2}{N^2}\sum_{i<k} (k-i) \sum_{j=i}^k\left\Vert Z_{j+1} - Z_j \right\Vert_2^2 \\
& \leq \frac{2}{N^2} \binom{N}{2} (N-1) \sum_{j=0}^{N-1}\left\Vert Z_{j+1} - Z_j \right\Vert_2^2 \\
& = \frac{(N-1)^2}{N}L_1
\end{align}

\end{proof}







\section{Detailed Rank Tables}\label{sec:rank_tables}
\begin{figure}[tbh]
    \centering
    \includegraphics[width=0.9\textwidth]{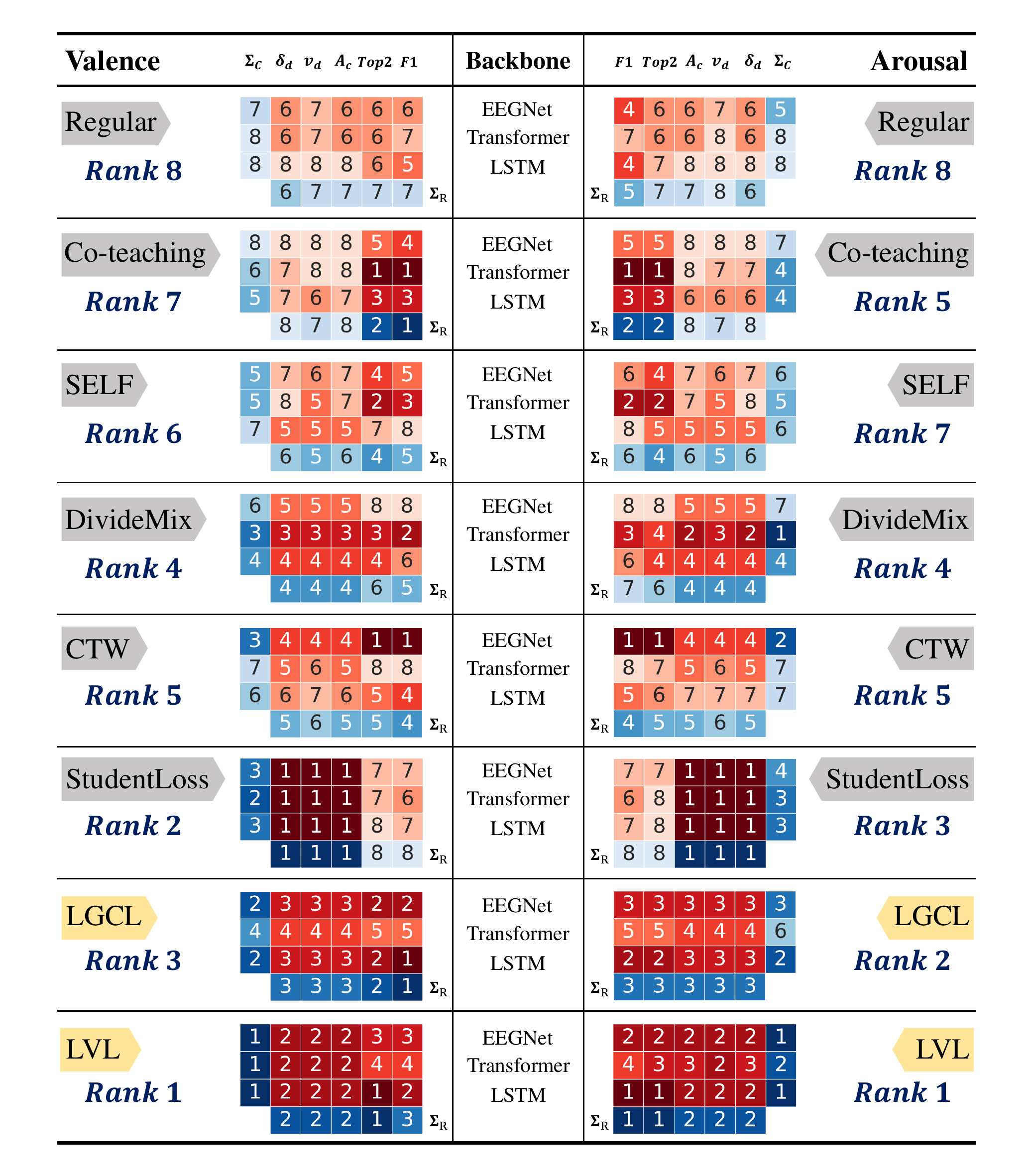}
    \caption{Summary of rank tables from different comparisons under 20\% label noise (DREAMER). Darker colors indicate higher ranks (better results).}
    \label{fig:summary20}
\end{figure}

\begin{figure}
    \centering
    \includegraphics[width=0.9\textwidth]{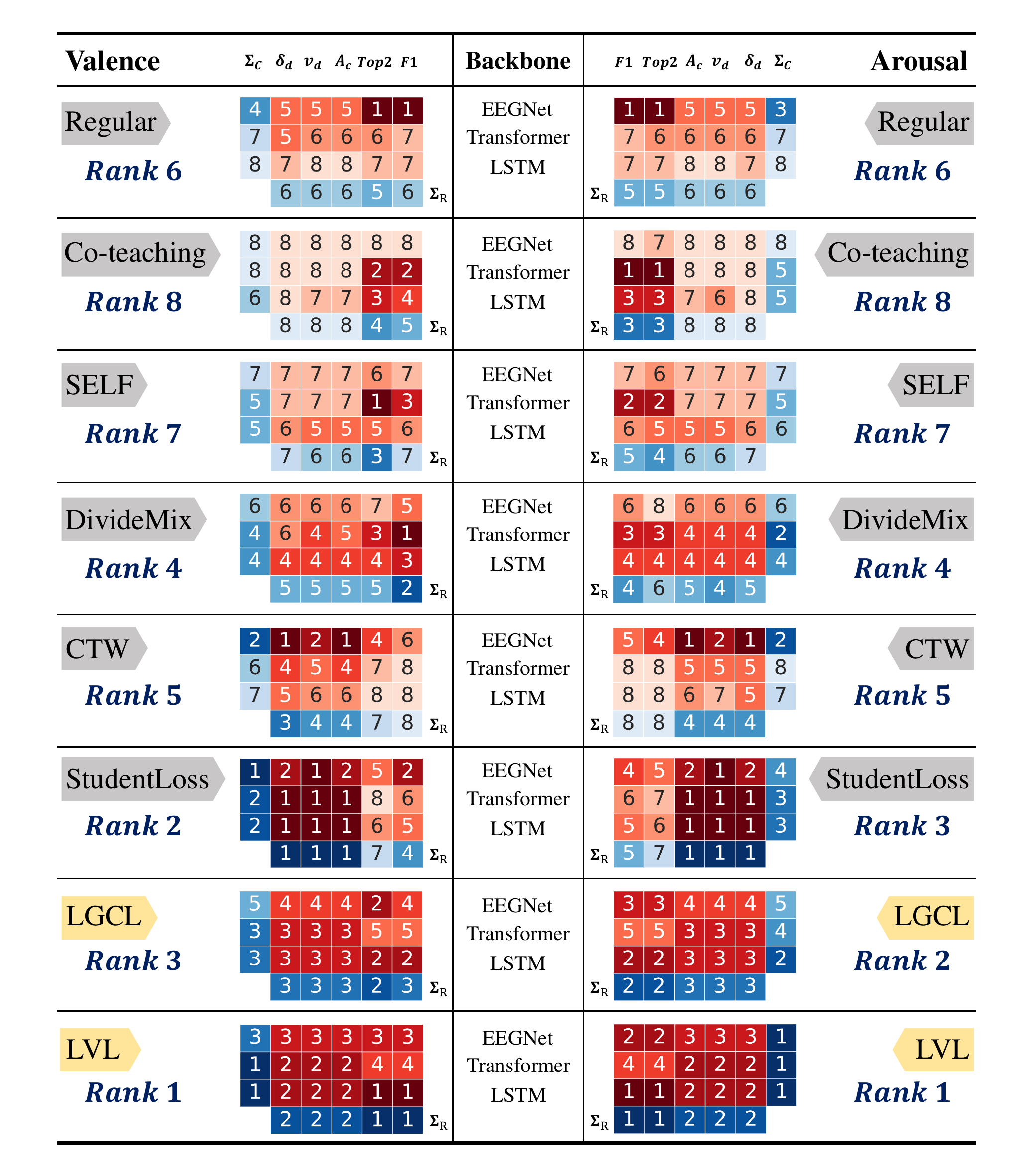}
    \caption{Summary of rank tables from different comparisons under 40\% label noise (DREAMER). Darker colors indicate higher ranks (better results).}
    \label{fig:summary40}
\end{figure}

\begin{figure}
    \centering
    \includegraphics[width=0.9\textwidth]{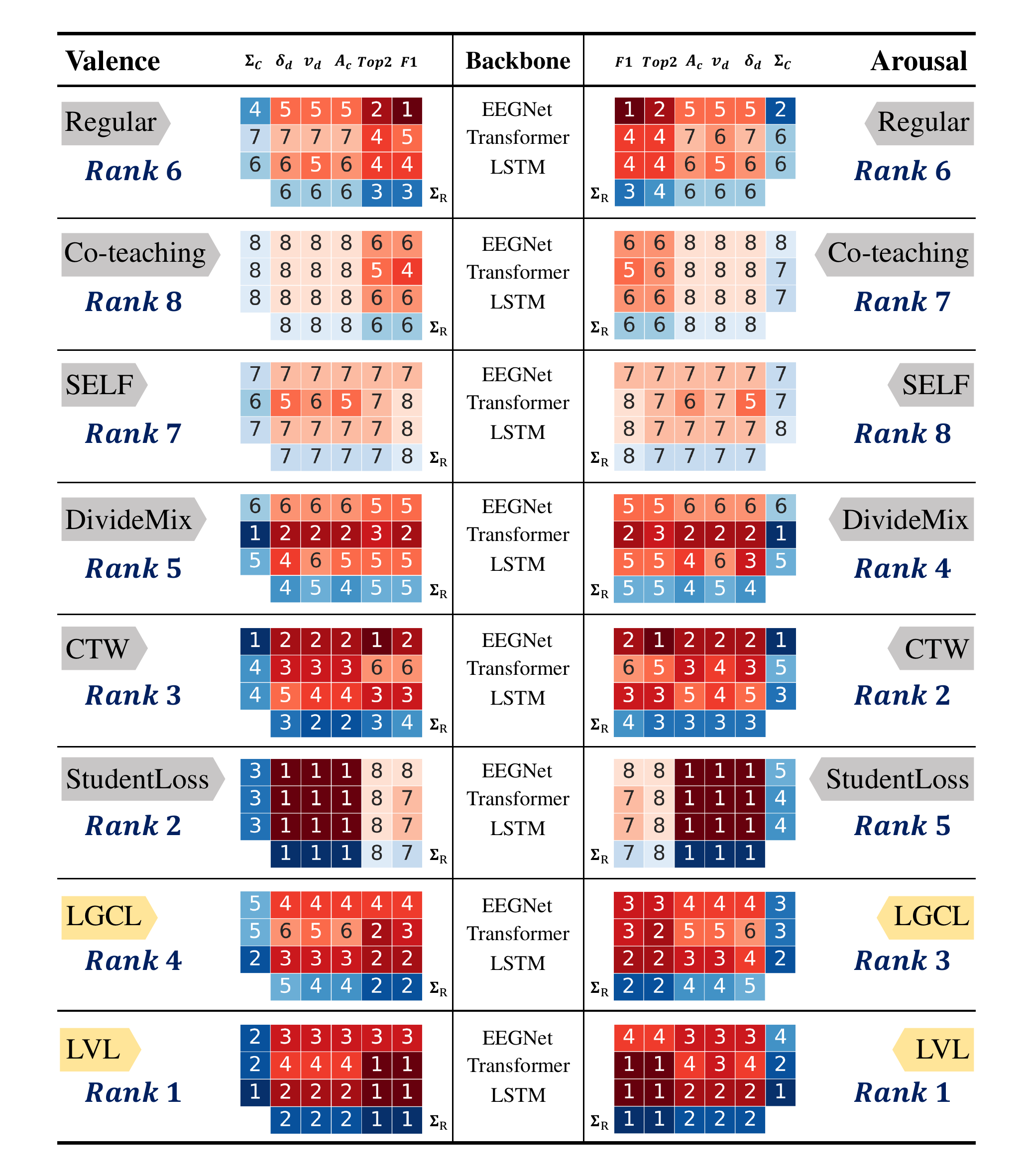}
    \caption{Summary of rank tables from different comparisons under 20\% label noise (DEAP). Darker colors indicate higher ranks (better results).}
    \label{fig:summary20_deap}
\end{figure}

\begin{figure}
    \centering
    \includegraphics[width=0.9\textwidth]{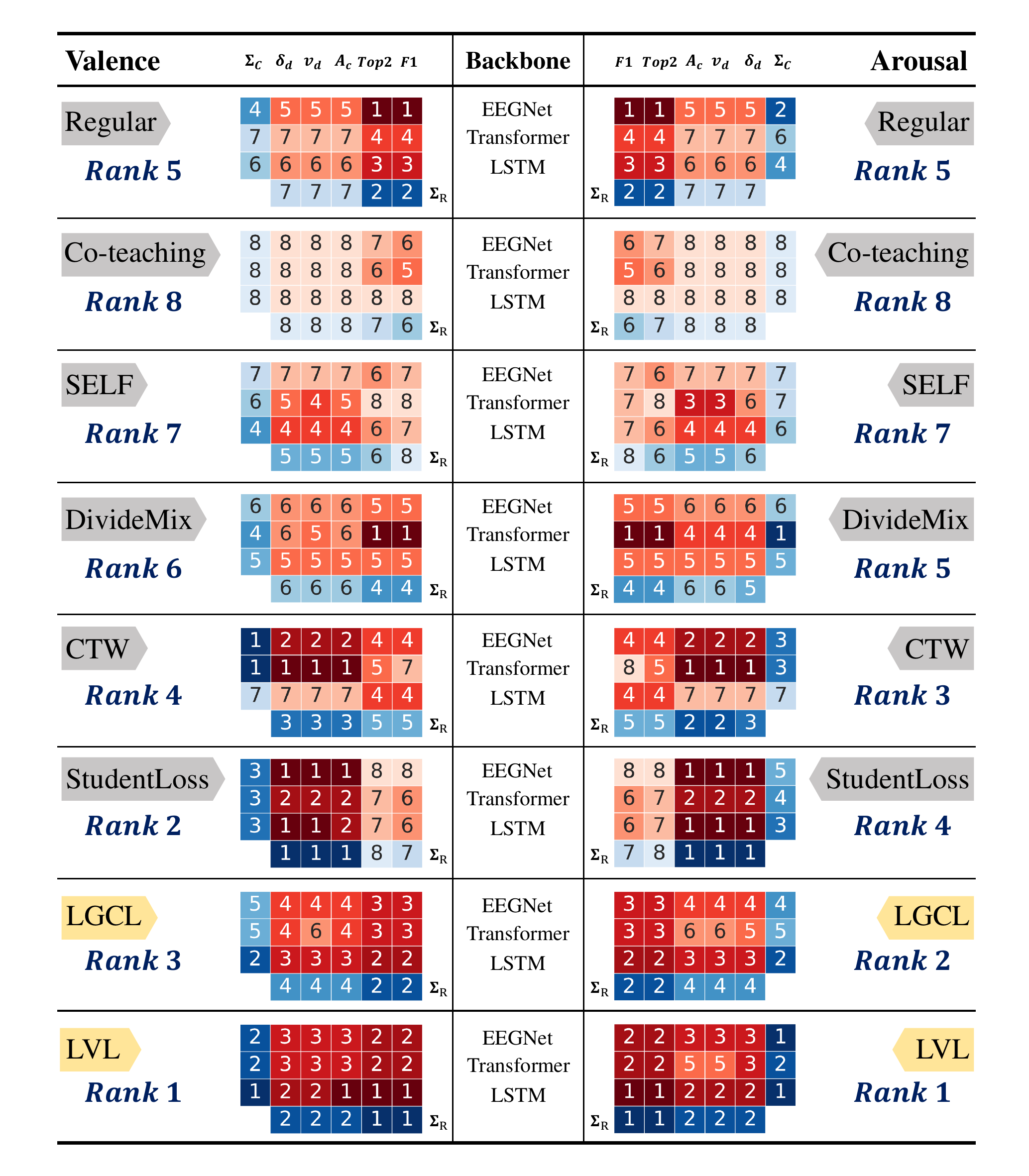}
    \caption{Summary of rank tables from different comparisons under 40\% label noise (DEAP). Darker colors indicate higher ranks (better results).}
    \label{fig:summary40_deap}
\end{figure}

Figure~\ref{fig:summary20} and Figure~\ref{fig:summary40} present detailed rank tables on DREAMER that compare eight training methods and multiple network backbones under 20\% and 40\% label noise, respectively. Figure~\ref{fig:summary20_deap} and Figure~\ref{fig:summary40_deap} are results from DEAP. Each cell in these tables display the corresponding rank (from 1 to 8) under the specified condition. For example, in the ``Regular'' cell of the \emph{Valence} column under 40\%  label noise, the number 4 in the first row and $\delta_d$ column (highlighted in a warm color) indicates that the ``Regular'' method is fourth in $\delta_d$ when EEGNet serves as the backbone. The full ranking for $\delta_d$ with EEGNet, ordered from best to worst, is: CTW, StudentLoss, LVL, LGCL, Regular, DivideMix, SELF, and Co-teaching.

Columns labeled $\Sigma_C$ (shaded in cool colors) show each method's aggregated rank across different metrics for a given backbone: lower numbers denote better overall performance on that backbone. Likewise, the bottom row ($\Sigma_R$), also shaded in cool colors, reports each method's aggregated rank across all backbones for a particular metric, reflecting the method's generalizability. An overall aggregated rank is displayed alongside each method's section of the table, combining all backbones and metrics. As a visual aid, darker colors correspond to better relative performance. 


In our experiments, we have several interesting observations. DivideMix, the most intricate LNL method in our benchmark, performs particularly well with the transformer-based model in terms of F1 metric for both datasets. Meanwhile, Co-teaching with a transformer shows a mixed pattern; it excels in quantitative measures (F1 and Top-2 accuracy), yet falters in qualitative metrics ($\delta_d, v_d, A_c$). Conversely, CTW paired with EEGNet achieves satisfactory qualitative performance but yields weaker results in quantitative metrics. While StudentLoss achieves impressive results on qualitative metrics, its performance on quantitative metrics is relatively poor. Its skewness among these five metrics is the most noticeable. Notably, LVL and LGCL exhibit weaker quantitative scores (e.g., F1 or Top-2 accuracy) when paired with a transformer on the DREAMER dataset, despite otherwise moderate standings among state-of-the-art methods. 

\end{document}